# An Agentic AI System for Multi-Framework Communication Coding

Bohao Yang[1], BS · Rui Yang[2,3], MS · Joshua M. Biro[4], PhD · Haoyuan Wang[5], MS · Jessica L. Handley[4], MA · Brianna N. Richardson[6], BS · Sophia Bessias[7], MPH · Nicoleta Economou-Zavlanos[1], PhD · Armando D. Bedoya[1,8], MD · Monica Agrawal[1], PhD · Michael M. Zavlanos[9], PhD · Anand Chowdhury[8], MD· Raj M. Ratwani[4, 10], PhD · Kai Sun[8], MD · Kathryn I. Pollak [6,11,#], PhD · Michael J. Pencina[1,#], PhD · Chuan Hong[1,#*] , PhD

1 Department of Biostatistics and Bioinformatics, Duke School of Medicine, Durham, NC, USA

2 Department of Population Health Sciences, Weill Cornell Medicine, New York, NY, USA

3 Systems Engineering, Cornell University, Ithaca, NY, USA

4 Medstar Health National Center for Human Factors in Healthcare, Washington, DC, USA

5 Applied Mathematics and Computational Science, University of Pennsylvania, PA, USA

6 Cancer Prevention and Control Research Program, Duke Cancer Institute, Durham, NC, USA

7 AI Evaluation & Governance, Duke Health, Durham, NC, USA

8 Department of Medicine, Duke School of Medicine, Durham, NC, USA, Durham, NC, USA

9 Department of Mechanical Engineering & Materials Science, Duke University, Durham, NC, USA

[10] School of Medicine, Georgetown University, Washington, DC, USA

[11] Department of Population Health Sciences, Duke School of Medicine, Durham, NC, USA

#Kathryn I. Pollak, Michael J. Pencina and Chuan Hong are equally contributed

Correspondence to: Chuan Hong. Email: chuan.hong@duke.edu

## ABSTRACT

Clinical communication is central to patient outcomes, yet large-scale annotation of patient–provider conversations remains labor-intensive, inconsistent, and difficult to scale. Existing large language model (LLM)–based approaches typically rely on single-task systems with limited robustness across communication frameworks and clinical domains. In this study, we developed MOSAIC, a Multi-framewOrk Structured Agentic AI system for scalable Clinical Communication coding. MOSAIC combines dynamically few-shot prompting, retrieval-augmented generation (RAG), and a LangGraph-based multi-agent workflow for planning, annotation, and verification. Performance was evaluated against gold-standard annotations produced by trained human coders. MOSAIC was developed on 54 annotated transcripts and evaluated on 120 transcripts spanning rheumatology and obstetrics/gynecology (OB/GYN). On the evaluation set, MOSAIC achieved an overall weighted F1 score of 95.6%. Performance was highest in rheumatology (F1 = 95.9%) and strongest for bias and patient behaviors categories (F1 = 96.4% and 96.1%, respectively). These results demonstrate the potential of multi-framework AI systems for reliable clinical communication analysis.

## INTRODUCTION

Clinician-patient communication quality is a modifiable, measurable driver of clinical outcomes, with specific behaviors (e.g., open-question rate, reflection-to-question ratio, interruptions, response latency, and empathic uptake) shaping trust, adherence, and fairness-sensitive endpoints.[1,2] Evidence-based communication techniques have been found to help patients change behaviors,[3] and communication-centered workflows that surface adherence barriers enable targeted problem-solving in chronic care.[4] These practices yield downstream benefits, such as improved blood pressure control and lower readmissions with stronger discharge communication.[5,6]

To improve communication quality, clinicians and patients need feedback and practice, not just good intentions. That feedback depends on communication coding to label key verbal and nonverbal behaviors (from utterances to whole visits), transforming real-world interactions into analyzable data.[7,8] Using explicit codebooks and decision rules, human coders annotate behavioral presence, frequency, quality, and timing to derive metrics such as rates, durations, and transitions. Example event-level labels include *empathic opportunities* and *empathic responses*,[9,10] *open-ended questions* and *reflective statements*,[1,11] *interruptions*,[12] and *shared decision-making moves*.[13] Global scales summarize relational climate (e.g., *respect*, *warmth*, *attentiveness*, *flow*).[7,14] The resulting time-stamped sequences support research, development, fidelity audits, and quality improvement.[15,16]

Recently, as the volume of recorded patient–provider interactions grow, the need for scalable, reliable coding has become increasingly urgent. For example, a 10-week pilot study reported AI scribe use in over 300,000 patient encounters involving thousands of physicians,[17] illustrating how AI scribing fuels both the demand and the opportunity for automated approaches to communication coding. Scaling communication coding is operationally challenging: manual multi-framework annotation requires reconciling overlapping codebooks, increasing cognitive load and cost, while double-coding and adjudication reduce consistency, particularly for rare or ambiguous labels and after codebook updates. This motivates tools that calibrate uncertainty, route low-confidence cases to humans, and learn from adjudication feedback.

LLMs can generate structured annotations from text, but most existing approaches rely on single-pass prompting without feedback or correction. Embedding LLMs in agentic architectures mitigates this by introducing tools, memory, and goal-directed loops.[18] Extending further, multi-agent systems can coordinate specialized LLM-agents across subtasks, an approach well-suited to communication coding, where overlapping frameworks and rare behaviors must be managed in parallel.[19,20] However, multi-agent communication coding system remains unexplored due to challenges in framework selection, conflict resolution, and uncertainty calibration.[21,22]

In this study, we developed MOSAIC, a Multi-framewOrk Structured Agentic AI system for Clinical Communication. Designed to meet the growing demand for scalable and reliable communication coding, MOSAIC coordinates multiple specialized AI agents to annotate clinical conversations across overlapping frameworks, such as empathy, behavior change, and bias. Unlike existing single-task approaches, MOSAIC dynamically selects the coding frameworks and reconciles conflicting labels. It also adapts through iterative refinement of prompts and example libraries based on adjudicated feedback, enabling more consistent and transparent annotation. By combining automation with human oversight, MOSAIC offers a practical and extensible solution for turning large volumes of patient–provider dialogue into structured, actionable insights.

## RESULTS

### Overview

As illustrated in Fig. 1, we developed MOSAIC, a structured multi-agent system for multi-framework clinical communication coding and applied it to clinician-patient transcripts from rheumatology and OB/GYN encounters. The broader workflow includes upstream preprocessing and transcript segmentation (Supplementary Fig. S1), while the core annotation engine combines codebook-grounded RAG within the Annotation Agent (Supplementary Fig. S2) with a dynamic example library that supports verification-informed in-context prompting refinement (Supplementary Fig. S3). For each requested framework, MOSAIC generates independent annotations, adjudicates disagreements, and evaluates final outputs against gold-standard annotations produced by trained human coders. Analyses focused on annotation accuracy, robustness across clinical domains and coding frameworks, and the contribution of individual system components.

The Results section is organized as follows. First, we describe the characteristics of the development and evaluation datasets and summarize the distribution of annotations across communication codebooks. Next, we evaluate MOSAIC's annotation performance, reporting overall accuracy as well as codebook-, department-, and transcript-level variability. We also compare MOSAIC against baseline and ablated variants to quantify the contribution of key system components, including RAG, dynamic prompting, dual coders and verification. We then present an explainability and error analysis section, examining common failure modes such as hallucinated or unsupported labels, ambiguous rule application, boundary or sentence-level misalignment, and low-confidence disagreements between coders, with representative examples and model-generated rationales describing the supporting transcript evidence and retrieved rules for assigned codes. Sensitivity analyses assess robustness across parameter settings, and a cost and efficiency analysis evaluates scalability relative to human annotation workflows. Finally, we present the MOSAIC user interface to illustrate usability, transparency, and end-to-end deployment.

**Dataset Characteristics**

The development dataset comprised 54 gold-standard annotated patient-provider transcripts (Table 1a). Clinical domains included rheumatology (n = 44) and OB/GYN (n = 10). Together, the transcripts contained approximately 22 hours and 47 minutes of speech, with an average transcript length of 24.86 minutes (SD = 13.94). On average, each transcript contained 303.04 speaker turns (median = 283.00, IQR 206.50–408.00), 361.84 sentences (median = 339.00, IQR 248.50–495.50), and 2756.11 words (median = 2629.00, IQR 1776.50–3575.50). Across these transcripts, human annotations identified 409 speaker turns with communication labels and 16,258 turns annotated as None. Labeled turns occurred at a mean of 7.44 per transcript (SD = 5.50; median = 7.00, IQR 3.50–9.50).

The evaluation dataset included 120 gold-standard annotated transcripts (rheumatology = 103; OB/GYN = 17) (Table 1b). These transcripts contained approximately 49 hours and 14 minutes of speech, with an average transcript length of 24.01 minutes (SD = 11.07). On average, each transcript contained 334.35 speaker turns (median = 319.00, IQR 241.50–432.00), 378.59 sentences (median = 354.00, IQR 265.00–484.50), and 3044.19 words (median = 2755.00, IQR 1951.00–3677.00). Human annotations identified 854 speaker turns with communication labels and 40,271 turns annotated as None. Labeled turns occurred at a mean of 6.94 per transcript (SD = 5.57; median = 5.00, IQR 3.00–9.00).

The distribution of annotations varied across codebooks in both the development and evaluation sets. Patient Behavior and WISER contributed the largest share of annotations in both datasets, followed by Bias and Global, while Intervention was less frequent. SDOH & Weight-related communication was included only in the development set. Together, these codebooks captured complementary dimensions of clinical communication, including clinician empathy and behaviors, patient communication behaviors, encounter-level relational quality, behavior change strategies, bias-related cues, and social determinants of health or weight-related communication (Supplementary Table S3).

**Qualitative Evaluation of Plan Agent**

Before reporting annotation performance, we first assess the Plan Agent's routing behavior, as correct framework selection is a prerequisite for downstream annotation accuracy. We qualitatively evaluated the Plan Agent's routing behavior (Supplementary eResult 1 and Table S4). It reliably mapped clear prompts to correct sub-agents, recognized synonyms (e.g., "empathy" → WISER), and handled malformed inputs.

**Annotation Performance**

*Overall and Codebook-Specific Performance.* Overall, MOSAIC achieved high annotation performance across all evaluation metrics. MOSAIC achieved an overall accuracy of 95.3% (95% CI: 94.8%–95.8%), precision of 96.1% (95.5%–96.6%), recall of 95.3% (94.8%–95.8%), and F1 of 95.6% (95.1%–96.1%) and the annotation performance varied by coding framework (Fig. 2a and Supplementary Table S6). Bias coding achieved the highest F1 (96.4%), followed by Patient Behavior (96.1%). Moderate performance was observed for WISER (95.2%) and Intervention (95.8%), while Global relational quality showed the lowest F1 (93.5%).

*Department-Specific Performance.* Performance remained strong across clinical domains, with rheumatology transcripts reaching an F1 of 95.9% (95.4%–96.4%), and OB/GYN transcripts showing moderately lower performance with an F1 of 94.1% (92.6%–95.5%) (Fig. 2b and Supplementary Table S6).

*Transcript-Level Variability.* Transcript-level performance variability revealed that MOSAIC consistently maintained high annotation quality across various clinical encounters, with a median F1 of 0.962 and tight interquartile ranges across all metrics (Fig. 2b and Supplementary Table

S11). Rheumatology transcripts (median F1 = 0.965) exhibited slightly higher median performance than OB/GYN (median F1 = 0.943), likely reflecting more structured dialogue patterns. A small number of transcripts showed performance dips, particularly in recall, corresponding to longer encounters or skewed label distributions.

*Comparison With Baselines and Ablation Variants.* We compared the full MOSAIC system with a series of baseline and ablation variants to quantify the contribution of each system component (Fig. 3). The single-agent baseline with codebook RAG achieved a weighted F1 of 0.895 (95% CI: 0.879, 0.911). Adding a codebook-specific prompt slightly improved performance to 0.904 (0.891, 0.922). The multi-agent configuration with a Plan Agent, an Update Agent, an Annotation Agent with single GPT-4o coder, verification, codebook RAG, and codebook-specific prompting achieved a comparable weighted F1 of 0.905 (0.890, 0.925), whereas replacing the GPT-4o coder with a GPT-5.2 coder reduced performance to 0.880 (0.867, 0.892). Adding an independent coder in Annotation Agent and an adjudication layer in Verification Agent substantially improved performance, with the GPT-4o adjudicator reaching 0.932 (0.923, 0.949) and the GPT-5.2 adjudicator reaching 0.940 (0.929, 0.952). The full MOSAIC system, which further incorporated dynamic few-shot prompting, achieved the best performance with a weighted F1 of 0.956 (0.951, 0.961), corresponding to an absolute improvement of 0.061 over the single-agent baseline. This pattern suggests that the largest gains came from the dual-coder design, adjudication, and dynamic few-shot refinement, rather than from the multi-agent architecture alone.

**Explainability and Error Analyses**

MOSAIC was designed to produce traceable coding decisions by linking assigned labels to retrieved codebook rules, supporting transcript evidence, and short verification rationales. High-confidence rationale-linked outputs were especially evident for labels capturing patient trust or reassurance signals, clinician use of open-ended questions, and patient assertive communication, suggesting that interpretable rule-to-label alignment was achievable in several stronger-performing frameworks. In the independent evaluation set, MOSAIC produced 43,199 correct final annotated labels. Final-output errors were primarily attributable to unsupported or hallucinated labels (1,253 cases) and missed labels (768 cases), whereas direct label confusion was uncommon (20 cases). Review-queue cases were mainly concentrated in alignment or formatting issues (1,722 cases), disagreements between a specific tag and no tag (772 cases), and disagreements involving multiple possible tags (519 cases), with missing rule support frequently cited (949 cases). These findings suggest that MOSAIC's human-in-the-loop workflow primarily surfaced cases that were ambiguous, weakly supported by retrieved rules, or difficult to resolve automatically. Residual errors were especially evident in encounter-level relational quality coding, where 65 review-queue rationales noted that a single utterance was insufficient for encounter-level judgment, suggesting that some communication constructs may require broader conversational context or multimodal cues beyond transcript text alone.

**Sensitivity Analyses**

We selected 0.3 as the default temperature for single GPT-4o coder and evaluated system performance across parameter configurations. Low temperatures (≤ 0.1) produced highly consistent but rigid outputs (e.g., 0.0, F1 = 90.2%). Higher temperatures (0.3, 0.5 and 0.7) occasionally captured greater nuance (F1 = 92.8%, 91.9% and 91.7% respectively) but tended to miss certain original content. These results show robust performance, with 0.3 yielding the best balance.

### Cost and Efficiency Analysis

We compared the time required for manual human annotation versus MOSAIC annotation. Human annotation required 60-90 minutes per 30-minute recording (approximately 20 minutes of clear speech), reflecting roughly 2-3 times the audio length and including double-coding of 10% of transcripts. For MOSAIC, we separated one-time or periodic setup time from transcript-level runtime. Setup time included retriever initialization, example-library embedding and index construction, startup warm-up, and codebook re-indexing when triggered by the Update Agent (4.84, 7.70, 3.14, and 0.10 seconds, respectively) (Supplementary Table S12). After setup had been completed, the average transcript-level runtime for full MOSAIC ranged from 4.91 to 8.21 minutes per transcript. Model-related token usage and API cost were estimated separately from runtime to distinguish computational expense from end-to-end workflow latency. For the full MOSAIC workflow using two GPT-4o coders and one GPT-5.2 adjudicator across 174 transcripts, Azure monitoring showed 42.01M input tokens and 10.77M output tokens, corresponding to approximately 241,437 input tokens, 61,897 output tokens, and an estimated $1.25 API cost per transcript, based on public per-token pricing and excluding cached-token discounts or institutional pricing adjustments (Supplementary Table S13).

### User Interface

To support end-to-end use, we developed a secure Gradio-based web interface (Supplementary Fig. S7) for transcript upload, pipeline execution, and annotation review. Users can select codebooks, adjust settings, and run MOSAIC with a single action. Outputs are presented in editable formats (.txt and .docx) with speaker turns and codes.

## DISCUSSION

In this study, we developed MOSAIC, a LangGraph-based multi-agent framework for annotating clinical communication. MOSAIC integrates a Plan Agent for workflow coordination, an Update Agent for codebook-aware retrieval updates, Annotation Agents using codebook and example-guided RAG, and a Verification Agent to ensure label consistency. Reliability is achieved through dynamic in-context prompting with an evolving example library, enabling structured and transparent annotation without model fine-tuning.

MOSAIC achieved strong overall performance (F1 = 95.6%), exceeding typical human inter-rater agreement reported in communication coding studies (75–90%) and matching or surpassing trained coder reliability for many annotation tasks.[1,23] Performance was highest in structured domains like rheumatology (F1 = 95.9%), while lower scores OB/GYN and Global codes reflected known challenges in human annotation. Performance gains were especially notable in Bias categories, where multi-agent coordination, codebook-guided retrieval, and dynamic in-context prompting improved recall and label balance. Sensitivity analyses showed a clear temperature trade-off of LLM coder: lower settings produced rigid consistency, while higher temperatures captured greater nuance at modest recall cost, with 0.3 offering the best balance. Runtime profiling showed that MOSAIC incurred modest one-time setup costs for retriever initialization, example-library indexing, startup warm-up, and codebook re-indexing, after which transcript-level runtime ranged from 4.91 to 8.21 minutes per transcript. Overall, MOSAIC achieves scalable, human-level annotation through structured, retrieval-informed orchestration.

To improve interpretability, we conducted sentence-level evaluation. Errors, especially in Global labels, reflected a granularity-completeness trade-off: despite lower performance than transcript-level runs, precision improved, highlighting the potential of audio cues and optimized chunking.

Building on these observations, qualitative patterns emerged (Table 2). In successful cases, MOSAIC matched or recovered human annotations, especially in the WISER and Intervention, identifying subtle empathic or reflective statements missed by coders. However, challenging cases in Global and Bias revealed limitations of text-only input, especially for distinctions (e.g., Attentive vs. Concerned and Guarded–Open) that depend on paralinguistic cues such as tone and speech tempo. These findings motivate multimodal extensions and position MOSAIC as a scalable annotator and diagnostic tool for voice-dependent ambiguity, supporting targeted human review and future audio-informed, human-in-the-loop integration.

Several broader limitations emerged. First, the current chunking strategy may not optimally preserve semantic coherence across conversational boundaries, potentially affecting recall. Second, the retrieval pipeline (MiniLM embeddings[24] and MedCPT reranking[25]) could likely be improved with newer domain-specific models. Third, the study was limited to rheumatology and OB/GYN, and a modest annotated sample size reflects the cost of gold-standard communication coding, which requires extensive clinician involvement and expert verification across overlapping codebooks. Although dynamic few-shot prompting alleviates data constraints without fine-tuning, larger datasets would further strengthen future evaluations. Fourth, some codebooks (e.g., SDOH, Weight) lacked sufficient gold-standard labels, and class imbalance, with most utterances labeled “None”, may bias metrics and limit sensitivity to rare but important signals. Global coding underperformed due to reliance on prosodic cues unavailable in text-only input. Performance also remains sensitive to codebook quality and LLM variability, while practical constraints include latency under multi-agent parallelism and cache management. Additionally, the dual-coder design relied on two independently executed instances of GPT-4o, with GPT-5.2 serving as the adjudicator. Although independent execution can yield different outputs, using the same underlying model for both coders may introduce correlated errors, particularly when ambiguous rules or transcript segments are systematically interpreted in the same way. A heterogeneous dual-coder configuration was not evaluated in the present study. Finally, race and ethnicity data were unavailable, precluding subgroup analyses. Although MOSAIC does not use race-based inputs, sociocultural and structural factors may influence communication behaviors. Future work should assess generalizability across diverse populations and incorporate fairness-aware methods.

Future work will target four areas: methodology, evaluation, generalization, and optimization. We will refine chunking and embedding strategies to preserve conversational coherence, explore reinforcement learning capabilities to enhance agent coordination and prompt adaptation, and incorporate multimodal inputs (e.g., tone, rhythm) for codes dependent on relational or affective signals such as Global and Bias. Evaluation efforts will assess whether clearer prompts improve categories like Intervention, test the Verification Agent’s effectiveness, and examine the use of LLMs as judges for converting annotations into decision-support insights. We will also compare same-model and heterogeneous dual-coder ensembles to assess whether model diversity reduces correlated errors and improves robustness, calibration, and disagreement-based adjudication. We also aim to generalize MOSAIC to additional clinical specialties, languages, and underrepresented codebooks (e.g., SDOH, Weight), thereby strengthening scalability and robustness under limited expert-labeled data.

In conclusion, MOSAIC demonstrates that an agentic, multi-agent framework can transform clinical communication coding into a scalable, accurate, and transparent process. By combining RAG, double coders and structured verification, and dynamic prompting, it improves consistency across diverse codebooks while preserving human-in-the-loop oversight. As AI scribes and ambient documentation tools increasingly generate growing volumes of clinical dialogue, systems like MOSAIC provide critical infrastructure to extract structured and trustworthy insights. This capability not only supports quality improvement, development, and research but also enables real-time feedback loops that promote shared decision making and foster high-quality care across clinical settings.

## METHODS

In this work, we developed MOSAIC, a structured, multi-agent system for multi-framework clinical communication coding, and applied it to clinician–patient transcripts including rheumatology and OB/GYN encounters. It was designed to support structured annotation across multiple communication frameworks through planning, codebook-grounded retrieval, independent dual coding, disagreement adjudication, and optional gold-standard verification. Here we detail the data sources, system architecture, and evaluation procedures used to develop and assess MOSAIC. We present the development and evaluation of MOSAIC in three parts: (1) data collection, including encounter recordings, gold-standard annotations, and codebooks (WISER, Global, Intervention, Patient Behavior, and Bias); (2) system architecture, featuring a LangGraph-based multi-agent framework, RAG, double coders and LLM as a judger, dynamic in-context prompting, and human-in-the-loop workflow; and (3) evaluation design, including benchmarks, metrics, explainability and error analyses, sensitivity analyses, and clinician feedback.

**Ethics and Consent**

This study was conducted in accordance with all relevant ethical regulations, including the Declaration of Helsinki, and was approved by the Duke University Health System Institutional Review Board (DUHS IRB; Pro00106131 and Pro00115358). Written and/or electronic informed consent was obtained from participating patients and clinicians using IRB-approved consent and e-consent forms.

**Data Collection**

We collected 174 patient–provider audio recordings across two clinical departments (rheumatology and OB/GYN), including 54 for development and 120 for evaluation. Transcripts were generated via an internal Ambient Digital Scribing (ADS) system from real clinical recordings and reviewed by clinicians for accuracy.[26] All transcripts were manually annotated by trained coders to create gold-standard labels. Gold-standard annotations were created through a structured, multi-coder annotation process. Each annotation team included 2–4 trained human coders who were trained using standardized coding instructions, scoring rubrics, and calibration examples. Coders proceeded to full annotation only after achieving inter-rater reliability of at least 80% across all codes during training. Discrepancies between coders were resolved through discussion and consensus.

## Codebook Configuration

Clinician-provided communication codebooks were reformatted for multi-agent use (Supplementary Table S2).[27] The *WISER Codebook* captures clinician behaviors and empathy; the *Global Codebook* assesses relational quality (e.g., flow, warmth, concern expression); and the *Intervention Codebook* supports fine-grained event-level coding of behavior change strategies (Ask, Advise, Assess, Assist, Arrange). Additional codebooks include *Patient Behaviors* (e.g., question-asking, assertiveness) and *Bias*, which flags both explicit and subtle cues, stereotyping, guardedness, rapport mismatch, and trust signals, reflecting clinical power dynamics.

## Proposed Multi-Agent Framework

The MOSAIC framework (Fig. 1 and Supplementary Table S1) employs a LangGraph-based[28] multi-agent architecture to support multi-framework clinical communication coding. The workflow begins with Preprocessing to segment and batch transcripts, which are then processed by the Core Agents (Plan, Update, Annotation, and Verification Agents) supported by RAG backbone.[29] The system securely interfaces with Duke's Azure OpenAI deployments to preserve patient privacy. In the current implementation, annotation is performed by two independent LLM coders, and disagreement adjudication within two coders is performed by a separate LLM-based judging layer, which in our primary experiments used GPT-5.2.[30]

*Preprocessing*. Transcripts from our internal ADS system[26] are preprocessed to ensure compatibility downstream MOSAIC annotation while preserving conversational context. Utterances are segmented with assigned speaker roles and timestamps. Transcripts are first partitioned into timestamp-delimited blocks in the raw transcript. Consecutive timestamp blocks are then concatenated into analysis batches until reaching a maximum length of 400 words per batch. This word-capped batching strategy controls prompt length while preserving temporally aligned and semantically coherent transcript segments (Supplementary Fig. S1).

*Plan Agent.* The Plan Agent serves as the workflow entry point and routing controller. It processed user-provided transcripts, codebooks, and task prompts; identifies the requested annotation frameworks; checks codebook availability; determines whether gold-standard labels were provided; and activates the appropriate downstream nodes. When users select specific frameworks, only the corresponding annotation agents are activated. When a new or revised codebook is detected, the Plan Agent routes the workflow to the Update Agent before annotation. When gold-standard labels are available, the Verification Agent is activated for development, calibration, or evaluation. For malformed, ambiguous, or unsupported inputs, the Plan Agent generates diagnostic warnings rather than proceeding with unsupported annotation.

*Update Agent.* The Update Agent handles new or revised codebooks by parsing their content and refreshing the FAISS[31] and retrieval index, ensuring that Annotation Agents access the most up-to-date rules. When a new codebook is provided, the updated framework is re-indexed and reloaded for the current annotation run; if no new codebook is detected, the agent proceeds with the existing configuration. If an incomplete codebook is provided, the Update Agent repairs it by aligning the new content with the original codebook structure. This process is strictly rule-based, rather than LLM-driven, to preserve traceability, minimize unintended modifications, and maintain the rigor required for clinical tasks.

*Annotation Agent and Double-Coder Architecture.* The Annotation Agent assigns communication codes using framework-specific sub-agents selected according to the requested task. Each specialized sub-agent corresponds to a clinician-defined communication-coding framework, including WISER, Global, Intervention, Patient Behavior, and Bias. For each transcript batch and requested framework, the Annotation Agent constructs a retrieval-augmented prompt and performs annotation using Azure OpenAI GPT-4o. The final annotation prompt was assembled from structured components: agent role definition and instructions, framework-specific constraints, retrieved codebook rules, dynamically selected few-shot examples, and the transcript batch to be annotated. This modular prompt structure constrained outputs to valid labels and reduced hallucinated or unsupported annotations (Supplementary Fig. S6). To emulate qualitative double-coding practice, the Annotation Agent generates two independent annotation outputs for each requested framework and transcript batch. The pipeline is executed independently twice, producing parallel outputs from two GPT-4o coders over the same input text. The outputs are then passed to the Verification Agent for disagreement detection, adjudication, and human-review flagging when needed.

*Codebook and Example-Library Retrieval.* The Codebook RAG module retrieves framework-specific rules from clinician-provided codebooks. Codebooks are reformatted with Markdown headers, segmented into rule-level chunks, embedded using MiniLM (all-MiniLM-L6-v2)[24] and indexed in FAISS. During annotation, the system retrieves rules via maximum marginal relevance (MMR),[32] reranks candidates using a MedCPT cross-encoder,[25] deduplicates redundant rules, and injects the top 10 rules into the annotation prompt (Supplementary Figs. S2 and S5). The Example Library RAG module supports dynamic few-shot prompting using verification-derived examples. Sentence-level examples are paired with local context and labeled as false negatives, false positives, true positives, or true negatives according to discrepancies between gold-standard annotations and model predictions. A hybrid BM25 and FAISS retrieval strategy selects relevant examples, with false negatives prioritized as “Focus” cases and false positives as “Avoid” cases. Up to five Focus and five Avoid examples are injected into each annotation prompt (Supplementary Fig. S3).

*Verification Agent.* The Verification Agent ensures output quality through a broader verification layer that includes adjudication, gold-standard comparison, uncertainty flagging, and structured feedback generation. An LLM adjudication sublayer (GPT-5.2) first compares outputs from two coders line by line using persistent line identifiers and resolves disagreements based on retrieved rules, producing a final label, confidence estimate, disagreement type, and rule-support note for each conflict case. In inference mode, low-confidence or ambiguous cases are flagged for human review. When gold-standard labels are available during development or calibration, the Verification Agent further compares final adjudicated outputs against reference annotations and computes structured evaluation metrics and mismatch reports to guide prompt and example refinement. In both settings, it generates structured feedback that supports ongoing system optimization.

*Feedback Optimization Loop.* The Feedback Optimization Loop converts verification metrics into prompt and annotation refinements. For each task, the Annotation Agent constructs a framework-specific prompt comprising: (1) system instructions specifying the active codebook, (2) validated annotation labels, (3) RAG-retrieved rule definitions and canonical examples, and (4) few-shot examples from the Example Library RAG, which retrieves validated annotation examples to guide in-context prompting, continuously updated via Verification Agent feedback (Fig. S3). Sentence-level instances paired with local context, human labels, and agent predictions are stored as candidate examples, including both supportive and contrastive cases. During the development phase, verification mismatches are converted into new example candidates, deduplicated, and appended to the example library; the example retriever is then refreshed for subsequent runs. For evaluation, the Example Library was initialized with pre-existing codebook examples and verified cases from the 54 development transcripts and frozen before processing the 120 held-out transcripts. Although few-shot examples were dynamically retrieved from this fixed library for each evaluation transcript, library updating and iterative prompt-refinement rounds were disabled. No evaluation labels, predictions, mismatches, adjudication outcomes, or examples from previously processed evaluation transcripts were incorporated into the library, and evaluation gold-standard labels were used exclusively for post hoc metric calculation.

**Evaluation and Performance Metrics**

MOSAIC was evaluated on 120 transcripts from rheumatology and OB/GYN contexts, with annotations derived from five codebooks (WISER, Global, Intervention, Patient Behavior, and Bias).

To assess the decision-making of the Plan Agent, we conducted a qualitative analysis of its routing performance. We examined its ability to activate appropriate annotation sub-agents based on user inputs and to flag malformed or ambiguous instructions. Transcripts and prompts spanning diverse routing scenarios, including synonymous, vague, and nonsensical inputs—were used to assess robustness and failure modes.

To assess the annotation performance of MOSAIC, we calculated accuracy, precision, recall, and F1 score at the transcript, clinical-domain, and codebook levels. At the transcript level, precision, recall, and F1 were computed as weighted averages, with each label weighted according to its frequency in the gold-standard annotations. This allowed more frequently occurring codes to contribute proportionally more to the overall estimates while still retaining information from less frequent codes. For clinical domain/ department comparisons (rheumatology vs. OB/GYN), metrics were computed per transcript and averaged within each subgroup. For codebook-specific performance, per-label metrics were aggregated across all transcripts associated with that codebook to provide a behavioral dimension–specific evaluation. We further estimated 95% confidence intervals using 10,000 bootstrap iterations for these estimates. Specifically, metric values were resampled with replacement 10,000 times, the mean was recomputed for each bootstrap replicate, and the 2.5th and 97.5th percentiles of the bootstrap distribution were used as the lower and upper confidence bounds, respectively.

To assess the contribution of system components, we compared the full MOSAIC system with a series of baseline and ablation variants on the 120-transcript evaluation set. The first baseline used a single LLM with codebook RAG only. A second single-agent variant added codebook-specific prompting. We then evaluated a multi-agent configuration that included the Plan Agent, rule-based Update Agent, Annotation Agent, Verification Agent, codebook RAG, and codebook-specific prompting, but used only a single coder. To examine model effects, we compared this single-coder multi-agent configuration using either a GPT-4o coder or a GPT-5.2 coder. We next evaluated whether adding an independent second coder and an adjudication layer improved performance, comparing adjudication with GPT-4o versus GPT-5.2. Finally, the full MOSAIC system incorporated all components, including multi-agent coordination, codebook RAG, two independent GPT-4o coders, GPT-5.2 adjudication, and dynamic few-shot prompting. This ablation design allowed us to distinguish the contribution of codebook-specific prompting, multi-agent orchestration, model choice, dual-coder adjudication, and dynamic few-shot refinement.

**Explainability and Error Analysis**

Explainability and error analysis were based on structured outputs from MOSAIC, including matched and mismatched annotations, disagreement reports, adjudicated labels, verification rationales, and review-queue records. Explainability was assessed by examining whether assigned labels could be traced to retrieved codebook rules, supporting transcript evidence, and model-generated rationales. Error analysis compared MOSAIC outputs against gold-standard human annotations and categorized discrepancies as unsupported or hallucinated labels, missed labels, label-confusion errors, or other residual mismatches. We further analyzed internal disagreement patterns between two independent coders, adjudication confidence scores, and review-queue cases to identify ambiguous predictions and characterize failure modes. Representative examples were selected to illustrate both successful rationale-linked predictions and recurrent error patterns.

**Sensitivity Analysis**

To assess robustness and design trade-offs, we varied key parameters in a series of experiments. We tested temperatures (0.0, 0.1, 0.3, 0.5, 0.7) for the single GPT-4o coder, balancing stability at lower settings against potentially richer outputs at higher ones.

**Cost and Efficiency Analysis**

To assess potential gains in scalability and efficiency, we compared the time required for manual human annotation versus automated MOSAIC annotation. For human coding, we recorded the average time required to annotate transcripts across varying codebooks and domains, based on input from trained coders involved in the gold-standard label generation. For MOSAIC, we separated one-time or periodic setup costs from transcript-level runtime. Setup costs included initial loading of codebook and example retrievers, example-library embedding and index construction, retriever warm-up at system startup, and codebook re-indexing when triggered by the Update Agent. Transcript-level runtime was measured across the full annotation process. Model-related token usage and API cost were estimated separately from runtime to distinguish computational expense from end-to-end workflow latency.

## DATA AVAILABILITY

The patient–provider conversation transcripts used in this study cannot be shared publicly due to privacy and institutional restrictions.

## CODE AVAILABILITY

All code used for model development, evaluation, and the MOSAIC system implementation is available at https://github.com/AutoEvaluation/MOSAIC

## ACKNOWLEDGMENTS

No funding was received for this research.

## AUTHOR CONTRIBUTIONS

MJP and CH conceptualized the study. BY, RY and CH led system design and evaluation. BY and RY implemented core technical components. BY, BR, KS, KIP conducted data curation and annotation pipeline development. KS and KIP advised on communication evaluation frameworks. BY, RY, JMB and AC contributed to system architecture and optimization. BR, KS, KIP, and MJP guided clinical integration and domain evaluation. All authors contributed to manuscript drafting and revision and approved the final version. BY and CH had full access to all the data in the study. CH takes responsibility for the decision to submit for publication.

## COMPETING INTERESTS

The authors declare no competing interests. R.M.R. is an Associate Editor of npj Digital Medicine but was not involved in the peer review or decision-making process for this manuscript. M.J.P. is employed by Optum outside the submitted work.

## TABLES

Table 1. Summary characteristics of patient-provider transcripts

a. Development dataset

| Characteristic | Value |
|---|---|
| Total transcripts | 54 |
| Clinical domains | Rheumatology (n = 44); Obstetrics/Gynecology (n = 10) |
| Total speech time | ~ 22h 47m |
| Average transcript length | ~ 24.86 minutes (SD ~ 13.94 min) |
| Sentences per transcript | Mean = 361.84 (SD = 140.57; range 130 - 629);<br>Median = 339 (IQR 248.50 - 495.50) |
| Words per transcript | Mean = 2756.11 (SD = 1289.85; range 835 - 5909);<br>Median = 2629 (IQR 1776.50 - 3575.50) |
| Speaker turns per transcript | Mean = 303.04 (SD = 137.39; range 76 - 557);<br>Median = 283 (IQR 206.50 - 408) |
| Gold-standard annotated transcripts | 54 |
| Codebooks applied | WISER, Global, Intervention (5As), Patient Behavior, Bias, SDOH & Weight |
| Total annotated turns with labels (from human annotation) | 409 speaker turns |
| Annotated turns with labels per transcript (from human annotation) | Mean = 7.44 (SD = 5.50; range 0 - 30);<br>Median = 7.00 (IQR 3.50 - 9.50) |
| Total turns annotated as None (from human annotation) | 16258 speaker turns |
| Turns annotated as None per transcript (from human annotation) | Mean = 295.50 (SD = 136.18; range 70 - 551);<br>Median = 279.00 (IQR 200.50 - 405.50) |

b. Evaluation dataset

| Characteristic | Value |
|---|---|
| Total transcripts | 120 |
| Clinical domains | Rheumatology (n = 103); Obstetrics/Gynecology (n = 17) |
| Total speech time | ~ 49h 14m |
| Average transcript length | ~ 24.01 min (SD ~11.07 min) |
| Sentences per transcript | Mean = 378.59 (SD = 152.93; range 102 - 809);<br>Median = 354 (IQR 265 - 484.50) |
| Words per transcript | Mean = 3044.19 (SD = 2074.71; range 784 - 16430);<br>Median = 2755 (IQR 1951 - 3677) |
| Speaker turns per transcript | Mean = 334.35 (SD = 136.91; range 54 - 670);<br>Median = 319 (IQR 241.50 - 432) |
| Gold-standard annotated transcripts | 120 |
| Codebooks applied | WISER, Global, Intervention (5As), Patient Behavior, Bias |
| Total annotated turns with labels (from human annotation) | 854 speaker turns |

| Characteristic | Value |
| --- | --- |
| Total transcripts | 120 |
| Clinical domains | Rheumatology (n = 103); Obstetrics/Gynecology (n = 17) |
| Total speech time | ~ 49h 14m |
| Average transcript length | ~ 24.01 min (SD ~11.07 min) |
| Sentences per transcript | Mean = 378.59 (SD = 152.93; range 102 - 809);<br>Median = 354 (IQR 265 - 484.50) |
| Words per transcript | Mean = 3044.19 (SD = 2074.71; range 784 - 16430);<br>Median = 2755 (IQR 1951 - 3677) |
| Speaker turns per transcript | Mean = 334.35 (SD = 136.91; range 54 - 670);<br>Median = 319 (IQR 241.50 - 432) |
| Gold-standard annotated transcripts | 120 |
| Codebooks applied | WISER, Global, Intervention (5As), Patient Behavior, Bias |
| Annotated turns with labels per transcript (from human annotation) | Mean = 6.94 (SD = 5.57; range 0 - 31);<br>Median = 5 (IQR 3 - 9) |
| Total turns annotated as None (from human annotation) | 40271 speaker turns |
| Turns annotated as None per transcript (from human annotation) | Mean = 327.41 (SD = 135.15; range 51 - 665);<br>Median = 314.00 (IQR 234.00 - 427.00) |

Table 2. Representative examples of single-label analysis

| Case Type | Codebook | Excerpt (Context) | Target Sentence | Human Annotation | Agent Annotation (MOSAIC) | Notes |
|---|---|---|---|---|---|---|
| Successful Case 1 | WISER | Clinician: “Do you have a boy or girl at home?” Patient: “Girl.” Clinician: “A girl.” | “A girl.” | — (missed) | Reflective Statement | MOSAIC correctly detected a Reflective Statement missed by the human coder. |
| Successful Case 2 | WISER | Patient: “Somebody I wanted to tell I was pregnant died before I could tell them.” Clinician: “Oh, I’m sure they know… they’re watching down on you.” | “Oh, I’m sure they know… they’re watching down on you.” | — (missed) | Reflective Statement | Correctly identified Empathic Statement missed by human. |
| Successful Case 3 | Intervention (5As) | Clinician: “Are you currently smoking any tobacco products?” Patient: “I’ve been doing patches off and on and still smoke, but much less now.” | “Are you currently smoking any tobacco products?” | ASK-START | ASK-START | Perfect alignment; correct identification of Ask phase initiation. |
| Successful Case 4 | Patient Behavior | Patient: “I struggle with headaches… I think some of it is from the plaquenil.” | “I struggle with headaches… I think some of it is from the plaquenil.” | Affective Response | Affective Response | Perfect agreement; correctly captured Affective Response. |
| Challenging Case 1 (Needs Voice) | Global | Clinician: “OK, sometimes you can try some vitamins… magnesium, melatonin, riboflavin.” | “OK, sometimes you can try some vitamins…” | Concerns: 4 | Attentive: 4 | Both labels plausible; distinction depends on tone and emphasis, highlighting the need for audio input. |
| Challenging Case 2 (Needs Voice) | Bias | Clinician: “So, a diagnostic test is where you go into where the baby is and you’re getting chromosomes… there’s a small chance of miscarriage.” | “So, a diagnostic test is where you go into where the baby is…” | Rushed: 4 | — (missed) | Speech tempo and intonation affect judgment; requires audio for accurate detection. |
| Challenging Case 3 (Needs Voice) | Bias | Clinician: “Do you think he would be willing to do that?” Patient: “Not really. No, I just have my fiancé…” | “Not really. No, I just have my fiancé…” | Guarded–Open: 4 | — (missed) | Requires prosodic cues (hesitation, tone) to differentiate guardedness vs openness. |

**FIGURE LEGENDS**

Figure 1. Overview of the MOSAIC workflow. Input data include user-provided dialogue transcripts and versioned clinician codebooks. Transcripts are preprocessed into structured, timestamp-aligned segments, then passed to the Agentic Core. The Plan Agent coordinates task routing and triggers the Update Agent when codebooks change. The Annotation Agent uses retrieval-augmented generation (RAG) to assign communication codes based on the codebooks via two independent GPT-4o coders. The Verification Agent resolves the conflicts between the two coders in Annotation Agent, compares outputs to gold-standard labels and provides feedback, which drives a feedback optimization loop. A dynamic Example Library RAG connects the Verification and Annotation Agents by incorporating adjudicated examples to iteratively refine few-shot prompts and improve annotation accuracy. This diagram was created using diagrams.net.

Figure 2. Annotation performance across codebooks and clinical domains. (a) *Overall and Codebook-Specific Annotation Performance.* Bar chart showing overall and codebook-specific accuracy, precision, recall, F1 scores and 95% Confidence Interval for five codebooks (WISER, Global, Intervention, Patient Behaviors, and Bias). (b) *Department-Specific Annotation Performance.* Distribution of MOSAIC's annotation performance across rheumatology and obstetrics/gynecology. This figure was created using Python.

Figure 3. *Performance Comparison of MOSAIC With Baselines and Ablation Variants.*

Grouped bar chart showing F1 scores for different system configurations, including the single-agent baseline (codebook RAG only, static instructions, no example-library retrieval, no verification, no dynamic prompting), multi-agent with codebook RAG alone, few-shot examples, dynamic in-context prompting and complete MOSAIC system. This figure was created using Python.

Figure 1. Overview of the MOSAIC workflow. Input data include user-provided dialogue transcripts and versioned clinician codebooks. Transcripts are preprocessed into structured, timestamp-aligned segments, then passed to the Agentic Core. The Plan Agent coordinates task routing and triggers the Update Agent when codebooks change. The Annotation Agent uses retrieval-augmented generation (RAG) to assign communication codes based on the codebooks via two independent GPT-4o coders. The Verification Agent resolves the conflicts between the two coders in Annotation Agent, compares outputs to gold-standard labels and provides feedback, which drives a feedback optimization loop. A dynamic Example Library RAG connects the Verification and Annotation Agents by incorporating adjudicated examples to iteratively refine few-shot prompts and improve annotation accuracy. This diagram was created using diagrams.net.

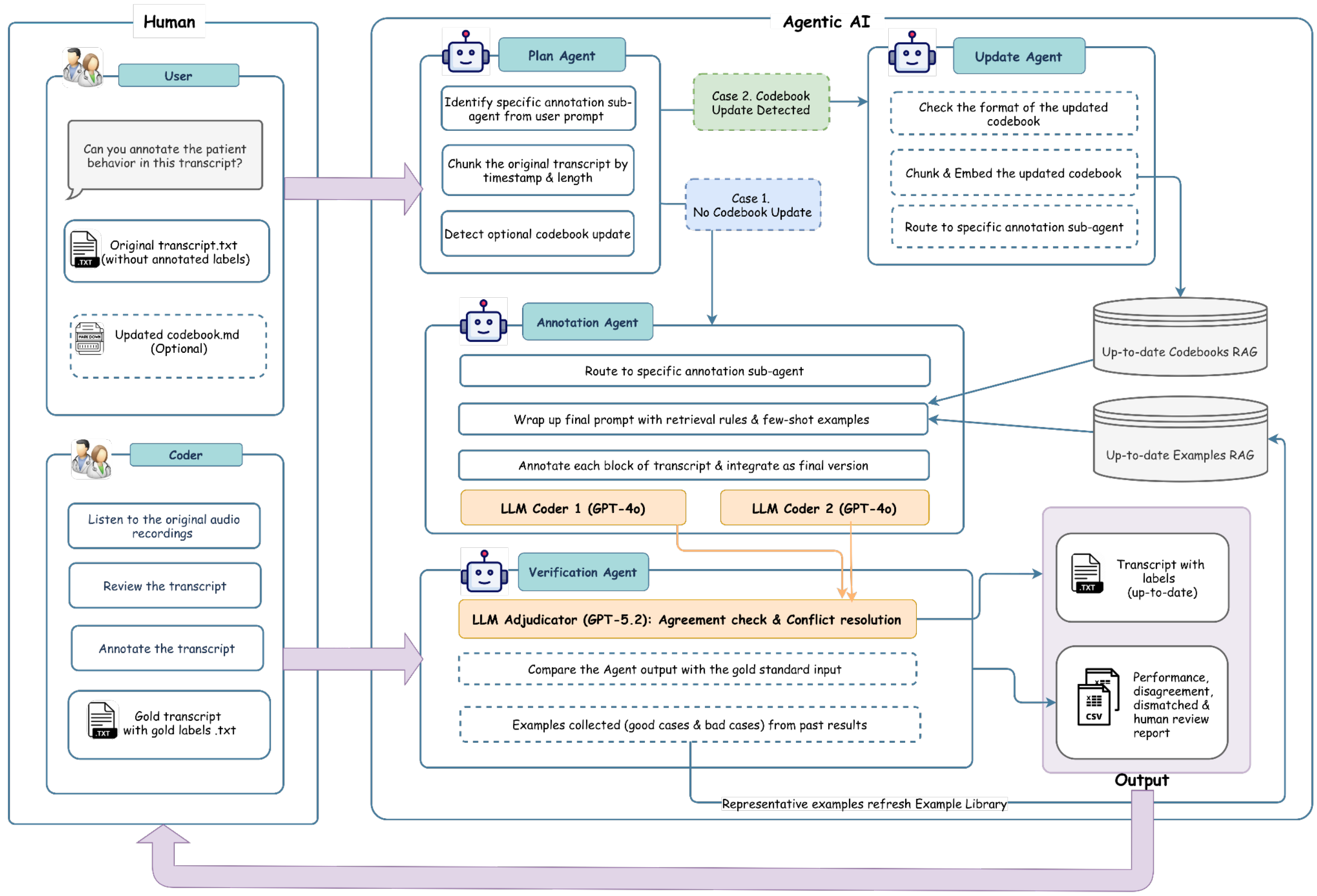

Figure 2. Annotation performance across codebooks and clinical domains. (a) *Overall and Codebook-Specific Annotation Performance.* Bar chart showing overall and codebook-specific accuracy, precision, recall, F1 scores and 95% Confidence Interval for five codebooks (WISER, Global, Intervention, Patient Behaviors, and Bias). (b) *Department-Specific Annotation Performance.* Distribution of MOSAIC's annotation performance across rheumatology and obstetrics/gynecology. This figure was created using Python.

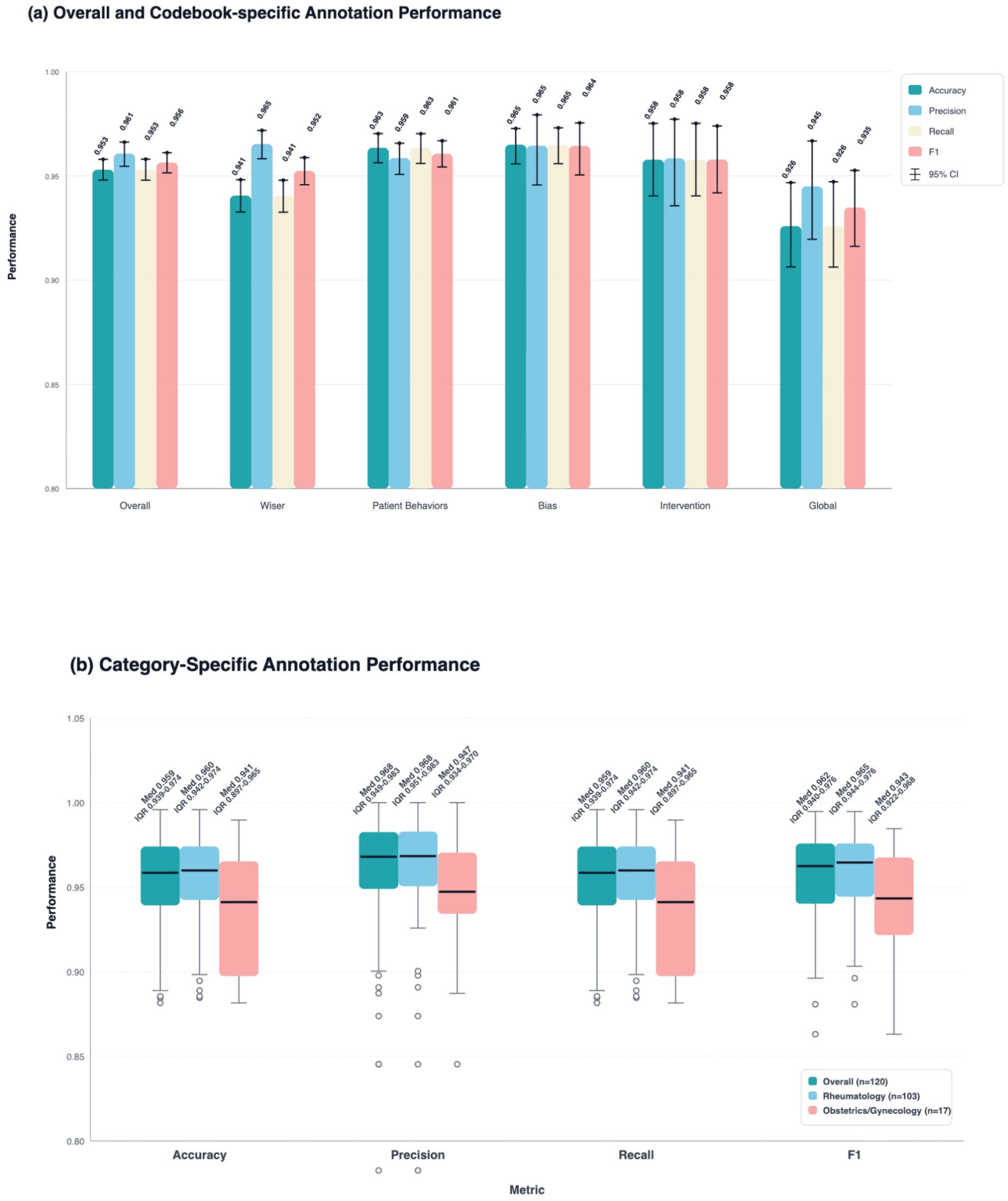

Figure 3. *Performance Comparison of MOSAIC With Baselines and Ablation Variants.*

Grouped bar chart showing F1 scores for different system configurations, including the single-agent baseline (codebook RAG only, static instructions, no example-library retrieval, no verification, no dynamic prompting), multi-agent with codebook RAG alone, few-shot examples, dynamic in-context prompting and complete MOSAIC system. This figure was created using Python.

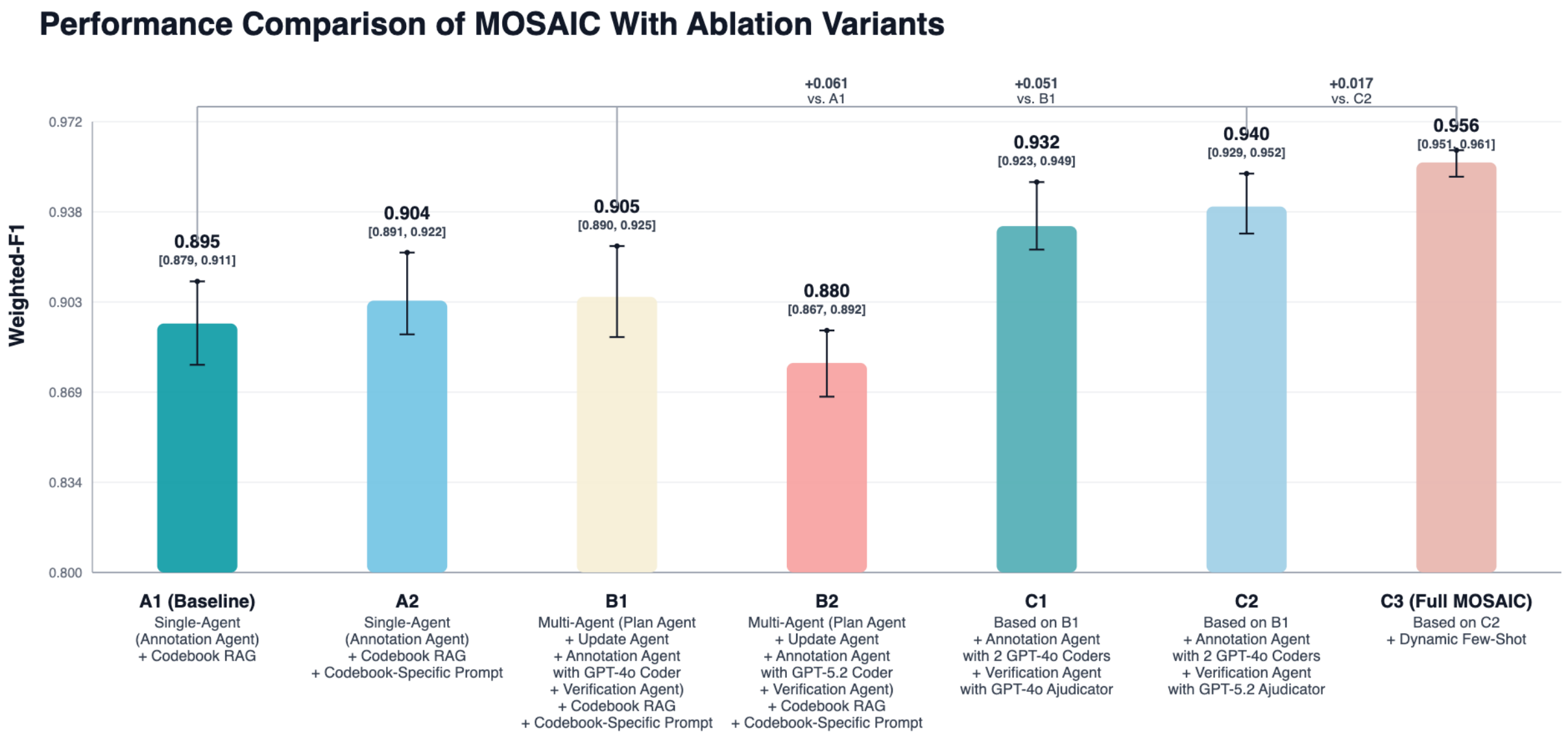

## Supplementary Materials

Supplement to: An Agentic AI System for Multi-Framework Communication Coding

## Table of contents

**Content** **Page**

## Supplementary Results

### eResult 1. Qualitative Evaluation of the Plan Agent

To evaluate the behavior of the Plan Agent, we conducted a qualitative analysis of its ability to activate the appropriate annotation sub-agent and to raise warnings when misrouting occurred. We examined transcripts where the Plan Agent flagged text segments and compared them against unflagged portions to understand the conditions under which its routing decisions were accurate or problematic. Representative prompts were used to illustrate performance across a range of conditions. In positive cases, the Plan Agent reliably mapped specialized categories (e.g., *Wiser*, *Patient Behavior*, *Intervention*, *Bias*, *Global*) to the correct sub-agents, even when prompts differed in wording, case, or synonyms. This shows robustness to expression variants and consistency with clear instructions. In negative cases, however, the agent struggled with ambiguous or underspecified prompts, such as "tone," "sentiment," "communication style," or "conversation flow", which were misrouted to inappropriate agents (e.g., confused among *Global, Wiser and Patient Behavior*) or flagged as invalid with no routing. Additionally, nonsense inputs ("asdfghjkl," "Test123") or negations ("run none") triggered useful warnings. These patterns highlight the Plan Agent's strengths in handling explicit, well-structured prompts, noisy inputs, synonyms, and requests involving multiple agents, where it generally produced consistent and accurate routing. However, its performance declined when the target (e.g. clinician and patient) of the instruction was vague or underspecified, leading to fewer correct matches or, in some cases, over-assignment of extra sub-agents compared to the intended standard. This underscores the need for improved disambiguation, synonym mapping, and fallback strategies in future iterations (Supplementary Table S4).

**eResult 2. Qualitative Analysis on Labelling Details**

At the label level, performance was highly uneven across annotation categories. The dominant None class showed strong performance (F1 = 0.976), indicating reliable identification of uncoded segments, whereas most low-frequency positive labels had substantially lower precision, recall, and F1. Moderate performance was observed for labels such as J (judgment; F1 = 0.118, support = 11), S (stereotyping; F1 = 0.076, support = 47), OE (open-ended question; F1 = 0.067, support = 120), ER (empathic response; F1 = 0.043, support = 52), EO (patient expresses emotion; F1 = 0.041, support = 28), AR (assertive response; F1 = 0.031, support = 215), AQ (patient asks questions; F1 = 0.024, support = 111), and RS (reflective statement; F1 = 0.019, support = 76). These labels often relied on subtle distinctions in conversational intent or interpersonal stance. For example, J could involve a clinician making a critical remark such as, *"Oh no, you're still eating that much fast food?"*, whereas S reflected assumptive language such as, *"I know this medication may be too expensive for you."* Likewise, OE required distinguishing genuinely open prompts such as, *"What worries you most about the treatment plan?"*, while RS depended on identifying reflective listening statements such as, *"So you're still struggling to keep food down."* Other labels, such as EO and ER, were also interactionally subtle: a patient statement like, *"I don't know how I'm going to handle this,"* constituted an empathic opportunity, whereas a clinician response such as, *"That sounds frightening,"* constituted an empathic response. Together, these findings suggest that MOSAIC performed well on the majority negative class but struggled with infrequent and behaviorally subtle labels, particularly those requiring contextual or interactional interpretation.

Detailed inspection of the poorest-performing labels further showed that many failures clustered in the Global and Bias frameworks. Several labels had zero true positives and F1 = 0, including I (interrupting), Attentive, 4, Concerns, 4, Flow, 4, Warmth, 4, Respect, 4, Rushed, 4, ES, and Attentive, 5, despite nonzero gold support. These labels generally captured higher-order interpersonal judgments rather than discrete lexical markers. For instance, I depended on recognizing moments in which a clinician cut off a patient mid-thought, whereas ES captured standalone empathic statements such as, *"This can be incredibly frustrating,"* even when no explicit emotional cue had immediately preceded it. Other poorly performing labels, such as Attentive, 4, Warmth, 4, and Respect, 4, reflected broader encounter-level impressions, including whether the clinician appeared engaged, sounded genuinely caring, or invited the patient's preferences in a respectful way. Similarly, labels such as Flow, 4 and Rushed, 4 depended on whether the conversation unfolded smoothly and at an unhurried pace, rather than on any single phrase. Many of these judgments likely depend on prosody, pacing, overlap, hesitation, emphasis, and other paralinguistic cues that are only partially preserved in text transcripts. Taken together, these error patterns suggest that transcript-only inference is insufficient for some nuanced communication constructs, and that incorporating audio information may be particularly important for improving performance on Global and Bias coding (Supplementary Table S5).

**eResult 3. Qualitative Analysis on Number of Sentences, Words, and Speaker Turns**

To examine whether transcript length influenced MOSAIC performance, we fitted three OLS regression models predicting transcript-level weighted F1 score from codebook type and one transcript-length measure: total number of sentences, total number of words, or total number of speaker turns. Across all three models, transcript length showed little evidence of a strong negative association with F1 performance. The coefficient for sentences per transcript was small and not statistically significant ($\beta = 2.28 \times 10^{-5}$, $p = 0.171$), and the coefficient for words per transcript was also small and did not reach statistical significance ($\beta = 2.03 \times 10^{-6}$, $p = 0.085$). Speaker turns showed a marginal positive association with F1 score ($\beta = 3.75 \times 10^{-5}$, $p = 0.050$), suggesting that longer or more interactive transcripts did not reduce model performance and may have provided slightly more contextual information for annotation. Across models, codebook type explained more variation in performance than transcript length. Using Bias as the reference category, Global coding consistently showed lower F1 scores across the sentence, word, and speaker-turn models, with coefficients around −0.031 and statistically significant p-values below 0.001. Wiser coding also showed a modest but significant reduction in F1 compared with Bias across all three models, with coefficients ranging from −0.0143 to −0.0156. In contrast, Intervention and Patient Behavior did not differ significantly from Bias after adjusting for transcript length.

Overall model fit was modest, with $R^2$ values ranging from 0.1139 to 0.1287 and adjusted $R^2$ values ranging from 0.0750 to 0.0905. Among the three transcript-length measures, the speaker-turn model had the slightly best fit based on $R^2$, adjusted $R^2$, AIC, and BIC. These findings suggest that MOSAIC's performance was relatively robust to variation in transcript length, while remaining more sensitive to differences in communication-codebook complexity, especially for Global and Wiser coding (Supplementary Fig. S8 and Table S6-S9).

**eResult 4. Subgroup/ Department Performance Differences and Implications for Multimodal Analysis**

To further contextualize these findings, we conducted additional statistical analyses to examine performance variability across clinical domains and transcript characteristics (Supplementary Table S11). These analyses revealed significantly lower F1 scores in OB/GYN encounters compared with rheumatology, with performance declines strongly associated with transcript length, sentence count, and speaker turn frequency. Such patterns suggest that OB/GYN conversations impose greater annotation difficulty due to richer emotional expression, interpersonal nuance, and rapid turn-taking, which are not fully represented in text-only transcripts. Notably, Global and bias-related codes exhibited the largest performance gaps, consistent with their reliance on affective and prosodic cues such as tone, emphasis, and speech rhythm. These results indicate that some low or near-zero F1 scores do not reflect model failure per se, but rather structural limitations of text-based annotation for communication behaviors that are inherently multimodal. Together, these findings empirically support the need for future speech–text fusion and multimodal modeling to more faithfully capture clinical communication dynamics, particularly in emotionally complex care settings.

**eResult 5. Explainability and Error Analysis**

MOSAIC was designed to improve interpretability by linking coding decisions to retrieved codebook rules, transcript evidence, and short adjudication or verification rationales. These explanations helped distinguish well-supported labels from uncertain predictions and identify cases requiring human review. High-confidence rationale-linked outputs suggested that interpretable rule-to-label alignment was feasible across several stronger-performing parts of the system (Supplementary Table S14-17).

Final-output error analysis showed that residual failures were mainly concentrated in annotations assigned without sufficient rule or transcript support (count = 1,708) and annotations that should have been assigned but were missed (count = 1,159). Label-substitution errors were uncommon (count = 27), suggesting that the main challenge was determining whether the transcript provided enough evidence for annotation rather than confusing one label with another. Unsupported annotations were more common in clinician communication and encounter-level relational coding, whereas missed annotations were concentrated in patient communication, clinician communication, and encounter-level relational coding.

Internal disagreement analysis showed that uncertainty was mainly driven by alignment or formatting problems, label-versus-no-label disagreements, and disagreements involving multiple plausible labels. Review-queue cases followed a similar pattern, with alignment or formatting problems (count = 1,722), label-versus-no-label disagreements (count = 772), and multi-label disagreements (count = 519) accounting for most flagged items. Rationale mining also showed frequent references to insufficient retrieved rule support (count = 949), as well as broader encounter-level judgment limitations (count = 48) and text-limited tonal or interpersonal cues (count = 80).

Representative examples suggested several recurring failure modes. Unsupported annotations often occurred when brief acknowledgments or minimal responses were over-interpreted as evidence of interpersonal or emotional communication. Missed annotations often involved subtle or context-dependent signals that were not explicit at the sentence level. Low-confidence review cases commonly reflected uncertainty about whether an utterance met the threshold for annotation.

Encounter-level relational coding showed the clearest limitation of transcript-only input. In 65 review-queue cases, rationales explicitly noted that a single sentence was insufficient to support an encounter-level judgment. These cases often required broader conversational context, sustained interactional evidence, or non-textual cues such as tone and pacing. This suggests that some remaining errors may reflect the structural limits of text-only coding and motivates future integration of audio-text information.

Overall, these analyses suggest that MOSAIC provides interpretable outputs, surfaces ambiguous cases for human review, and identifies residual errors that are clinically meaningful and structurally interpretable rather than random.

## Supplementary Tables

Table S1. Multi-Agent Structure Overview.

This table summarizes the core MOSAIC agents, their workflow functions, and whether each component uses LLM inference. MOSAIC is implemented as a LangGraph-based workflow with modular nodes, typed edges, shared state, and conditional routing. The Plan Agent routes transcripts, prompts, requested frameworks, updated codebooks, and optional gold-standard labels to downstream nodes. The rule-based Update Agent rebuilds retrieval resources when revised codebooks are uploaded. The Annotation Agent performs framework-specific coding using two independent LLM coders with codebook-guided retrieval. The Verification Agent adjudicates coder disagreements, generates final labels and review queues, and, when gold-standard labels are available, computes evaluation outputs and updates the dynamic example library. This structure supports traceable orchestration, modular execution, and optional human review.

| Agent | Function | LLM use |
|---|---|---|
| Plan Agent | Reads the transcript, user prompt, and requested annotation frameworks; writes a brief task plan to shared state and determines downstream routing conditions. | LLM-enabled planning |
| Update Agent | Rebuilds and reloads framework-specific codebook retrieval resources when a new codebook is uploaded. | Rule-based / deterministic, no LLM |
| Annotation Agent | Performs the core coding workload for each requested framework using two independent coder passes with codebook-guided retrieval. | LLM-based double coding; use GPT-4o for coders |
| Verification Agent | First adjudicates disagreements between the two independent coders, then optionally verifies final adjudicated outputs against gold labels, producing agreement statistics, review queues, and evaluation reports. | Hybrid: LLM-based adjudication using GPT-5.2/judging plus rule-based metric evaluation |

Table S2. Codebook and Representative Examples.

This table summarizes the clinician-defined communication codebooks used by MOSAIC and representative labels from each framework. The WISER captures empathic opportunities and clinician responses, including empathic responses, empathic statements, sorry statements, open-ended questions, reflective statements, and elicit questions. The Patient Behavior framework captures patient-initiated communication behaviors, including question-asking and assertive responses. The Global framework assesses encounter- or segment-level relational qualities, including flow, respect, warmth, attentiveness, and concern. The Intervention framework captures 5As behavior-change steps, including Ask, Advise, Assess, Assist, and Arrange. The Bias framework identifies explicit and subtle bias-related cues, including judgment, stereotyping, tailoring, interrupting, rapport-building, rapport mismatch, trust, and distrust. The SDOH and Weight framework was implemented in the interface to detect social determinants of health and weight-related discussions, but was not included in the primary evaluation because of limited relevant data.

| Codebook | Main Focus | Level of Detail | Unique Twist | Example |
|---|---|---|---|---|
| Global | Overall relational quality | Summary scores (1–5) | Good for simple benchmarking | Flow: [Flow: 4] — “Conversation flowed well, few interruptions.”<br>Respect: [Respect: 5] — “Clinician consistently asked permission before exam.” |
| Intervention (5A’s) | Specific behavior changes steps | Detailed event coding | Tracks guideline fidelity (e.g., smoking) | Asked: [ASK] — “Do you currently smoke?”<br>Assessed: [ASSESS] — “Would you like to quit in the next month?”<br>Assisted: [ASSIST w/ Sol]— “What barriers might get in the way of quitting?” |
| WISER | Empathy, | Captures empathy | | Open-ended Question: [OE] — “How’s your weekend?”<br>Empathic Opportunity: [EO] — “I’m really worried about paying for this medication.”<br>Reflective Statement: [RS] — “So you’re feeling unsure about the plan.” |
| Patient Behaviors | Patient agency | Captures patient behaviors | Tracks how patients push back or engage | Patient Asking: [AQ] — Patient: “Is this safe for the baby?”<br>Patient Assertive: [AR] — “I don’t want to take that medication right now.” |
| Bias | Trust, distrust, explicit bias | Combines behavior counts & global impression | Captures subtle power dynamics & explicit stereotyping | Judgement: [J] — “Oh no. You’re not still eating all that fast food, are you?”<br>Stereotyping: [S] — “Is the baby daddy still not helping out?”<br>Guarded-Open: [GO: 2] — “Patient sounded reserved, hesitant.”<br>Trust: [TP] — “So, you think I should keep taking this?”<br>Distrust: [D] — “But I really think we should do another test.” |

Table S3. Codebook-specific Characteristics of Patient-provider Transcripts

a. Development dataset

| Codebook | Number of transcripts | Sentences per transcript | Words per transcript | Speaker turns per transcript | Annotated turns with labels | Turns annotated as None |
|---|---|---|---|---|---|---|
| Wiser | 16 | Mean = 371.65 (SD = 145.00); Median = 330.00 (IQR = 260.00) | Mean = 2894.35 (SD = 1434.19); Median = 2629.00 (IQR = 1849.00) | Mean = 325.41 (SD = 145.57); Median = 300.00 (IQR = 253.00) | 124 | 5408 |
| Patient Behavior | 16 | Mean = 377.75 (SD = 147.60); Median = 356.00 (IQR = 270.00) | Mean = 2925.19 (SD = 1476.76); Median = 2727.00 (IQR = 2026.00) | Mean = 338.56 (SD = 139.72); Median = 312.00 (IQR = 239.25) | 131 | 5286 |
| Bias | 13 | Mean = 367.92 (SD = 102.71); Median = 353.00 (IQR = 192.00) | Mean = 2615.62 (SD = 835.37); Median = 2766.00 (IQR = 959.00) | Mean = 311.92 (SD = 103.08); Median = 304.00 (IQR = 188.00) | 60 | 3995 |
| Intervention (5As) | 2 | Mean = 149.00 (SD = 26.87); Median = 149.00 (IQR = 19.00) | Mean = 1061.50 (SD = 320.32); Median = 1061.50 (IQR = 226.50) | Mean = 89.50 (SD = 2.12); Median = 89.50 (IQR = 1.50) | 15 | 164 |
| Global | 5 | Mean = 279.80 (SD = 148.57); Median = 241.00 (IQR = 173.00) | Mean = 2321.20 (SD = 1099.31); Median = 2305.00 (IQR = 1077.00) | Mean = 156.60 (SD = 78.10); Median = 152.00 (IQR = 129.00) | 40 | 743 |
| SDOH & weight | 2 | Mean = 529.50 (SD = 40.31); Median = 529.50 (IQR = 28.50) | Mean = 3923.50 (SD = 673.87); Median = 3923.50 (IQR = 476.50) | Mean = 350.50 (SD = 53.03); Median = 350.50 (IQR = 37.50) | 39 | 662 |

b. Evaluation dataset

| Codebook | Number of transcripts | Sentences per transcript | Words per transcript | Speaker turns per transcript | Annotated turns with labels | Turns annotated as None |
|---|---|---|---|---|---|---|
| Wiser | 36 | Mean = 420.37 (SD = 154.25); Median = 381.50 (IQR = 208.25) | Mean = 3478.24 (SD = 2410.69); Median = 3037.00 (IQR = 1310.75) | Mean = 375.76 (SD = 124.87); Median = 334.00 (IQR = 201.75) | 281 | 13998 |
| Patient Behavior | 37 | Mean = 410.71 (SD = 153.02); Median = 370.50 (IQR = 216.75) | Mean = 3393.11 (SD = 2413.37); Median = 2994.50 (IQR = 1374.50) | Mean = 367.21 (SD = 122.96); Median = 326.50 (IQR = 194.75) | 302 | 13652 |
| Bias | 30 | Mean = 316.83 (SD = 121.18); Median = 315.00 (IQR = 181.00) | Mean = 2324.03 (SD = 1172.05); Median = 2181.00 (IQR = 1416.50) | Mean = 280.23 (SD = 119.18); Median = 290.50 (IQR = 180.50) | 124 | 8283 |
| Intervention (5As) | 5 | Mean = 197.40 (SD = 106.86); Median = 168.00 (IQR = 71.00) | Mean = 1437.00 (SD = 723.76); Median = 1288.00 (IQR = 318.00) | Mean = 114.20 (SD = 74.84); Median = 91.00 (IQR = 5.00) | 44 | 527 |
| Global | 12 | Mean = 374.50 (SD = 150.94); Median = 369.50 (IQR = 93.00) | Mean = 3034.83 (SD = 1261.11); Median = 3035.00 (IQR = 1939.00) | Mean = 326.17 (SD = 159.24); Median = 332.50 (IQR = 136.50) | 103 | 3811 |

Table S4. Representative Prompt Examples and Results of Qualitative Evaluation of the Plan Agent

| Prompt | Case | Expected Sub-Agent Routing | Sub-Agent Routing Decided by Update Agent | Verdict | Error Message |
|---|---|---|---|---|---|
| Run Bias and WISER | Basic positive | Bias, Wiser | Bias, Wiser | PASS | – |
| Please run Global | Basic positive | Global | Global | PASS | – |
| Run Patient Behavior and Intervention | Basic positive | Intervention, Patient Behavior | Intervention, Patient Behavior | PASS | – |
| Run all | Basic positive | Bias, Global, Intervention, Patient Behavior, SDOH & weight, Wiser | Bias, Global, Intervention, Patient Behavior, SDOH & weight, Wiser | PASS | – |
| run bias & wiser | Variant expression | Bias, Wiser | Bias, Wiser | PASS | – |
| RUN GLOBAL AND INTERVENTION | Variant expression | Global, Intervention | Global, Intervention | PASS | – |
| Annotate empathy and advice | Variant expression | Wiser, Intervention | Wiser, Intervention | PASS | – |
| Evaluate stigma and prejudice | Variant expression | Bias | Bias | PASS | – |
| Check SDOH factors and weight discussion | Variant expression | SDOH & weight | SDOH & weight | PASS | – |
| Only run Wiser | Single agent | Wiser | Wiser | PASS | – |
| Annotate just Bias | Single agent | Bias | Bias | PASS | – |
| Focus on patient behaviors | Single agent | Patient Behavior | Patient Behavior | PASS | – |
| Check obesity coding | Single agent | SDOH & weight | SDOH & weight | PASS | – |
| Run Wiser, Global, and Bias | Complex prompt | Bias, Global, Wiser | Bias, Global, Wiser | PASS | – |
| Please run intervention, bias, and sdoh | Complex prompt | Bias, Intervention, SDOH & weight | Bias, Intervention, SDOH & weight | PASS | – |
| Analyze communication style of the doctor | Ambiguous | Wiser | Global, Wiser | FAIL | – |
| Check tone and sentiment of patient response | Ambiguous | Patient Behavior | Wiser | FAIL | – |
| Do empathy coding for this transcript | Ambiguous | Wiser | Wiser | PASS | – |
| Look at overall dialogue quality | Ambiguous | Global | Global | PASS | – |
| Just check conversation length | Negative | – | – | PASS | No valid annotation agents found |
| Evaluate conversation flow | Ambiguous | Global | Global | PASS | – |
| Check doctor dominance in dialogue | Ambiguous | Bias | Bias | PASS | – |
| Identify unclear patient responses | Ambiguous | Patient Behavior | – | FAIL | No valid annotation agents found |
| Review if intervention was successful | Ambiguous | Intervention | Intervention | PASS | – |
| asdfghjkl | Extreme (nonsense) | – | – | PASS | No valid annotation agents found |
| Please annotate with all modules available | Extreme | Global, Wiser, Intervention, Bias, SDOH & weight, Patient Behavior | Global, Wiser, Intervention, Bias, SDOH & weight, Patient Behavior | PASS | – |
| Can you check social determinants, empathy, and overall coding? | Extreme | Global, Wiser, SDOH & weight | Global, Wiser, SDOH & weight | PASS | – |
| RUN EVERYTHING | Extreme | Global, Wiser, Intervention, Bias, SDOH & weight, Patient Behavior | Global, Wiser, Intervention, Bias, SDOH & weight, Patient Behavior | PASS | – |
| Please run none of the agents | Negative | – | – | PASS | No valid annotation agents found |
| Evaluate ALL aspects of patient-doctor communication | Extreme | Global, Wiser, Intervention, Bias, SDOH & weight, Patient Behavior | Global, Wiser, Intervention, Bias, SDOH & weight, Patient Behavior | PASS | – |
| Test123 !!! ??? | Extreme (nonsense) | – | – | PASS | No valid annotation agents found |

Table S5. Qualitative Analysis on Labelling Details

a. Detailed performance of good cases

| Label | Explanation | TP | FP | FN | TN | Total | Accuracy | Precision | Recall | F1 |
|---|---|---|---|---|---|---|---|---|---|---|
| None | no coded behavior | 41115 | 768 | 1253 | 63 | 43199 | 0.953 | 0.982 | 0.970 | 0.976 |
| J | judgmental language/tone | 1 | 5 | 10 | 2888 | 2904 | 0.995 | 0.167 | 0.091 | 0.118 |
| S | stereotyping | 3 | 29 | 44 | 13979 | 14055 | 0.995 | 0.094 | 0.064 | 0.076 |
| OE | open-ended question | 13 | 255 | 107 | 14710 | 15085 | 0.976 | 0.049 | 0.108 | 0.067 |
| ER | empathic response | 4 | 128 | 48 | 19496 | 19676 | 0.991 | 0.030 | 0.077 | 0.043 |
| EO | empathic opportunity | 3 | 114 | 25 | 11772 | 11914 | 0.988 | 0.026 | 0.107 | 0.041 |
| AR | assertive patient response | 6 | 167 | 209 | 15893 | 16275 | 0.977 | 0.035 | 0.028 | 0.031 |
| AQ | patient asks questions | 2 | 52 | 109 | 14398 | 14561 | 0.989 | 0.037 | 0.018 | 0.024 |
| RS | reflective listening statement | 2 | 132 | 74 | 14915 | 15123 | 0.986 | 0.015 | 0.026 | 0.019 |

b. Detailed performance of bad cases

| Label | Explanation | TP | FP | FN | TN | Total | Accuracy | Precision | Recall | F1 |
|---|---|---|---|---|---|---|---|---|---|---|
| I | interrupting the patient | 0 | 2 | 32 | 4415 | 4449 | 0.992 | 0.000 | 0.000 | 0.000 |
| Attentive, 4 | mostly attentive/listening well | 0 | 44 | 13 | 6071 | 6128 | 0.991 | 0.000 | 0.000 | 0.000 |
| Concerns, 4 | addresses concerns well | 0 | 10 | 12 | 3849 | 3871 | 0.994 | 0.000 | 0.000 | 0.000 |
| Flow, 4 | conversation flows well | 0 | 0 | 11 | 2501 | 2512 | 0.996 | 0.000 | 0.000 | 0.000 |
| Warmth, 4 | generally warm and caring | 0 | 26 | 11 | 5099 | 5136 | 0.993 | 0.000 | 0.000 | 0.000 |
| Respect, 4 | clearly respectful | 0 | 22 | 10 | 5698 | 5730 | 0.994 | 0.000 | 0.000 | 0.000 |
| Rushed, 4 | generally unrushed pace | 0 | 0 | 9 | 2310 | 2319 | 0.996 | 0.000 | 0.000 | 0.000 |
| ES | empathic statement | 0 | 25 | 7 | 7925 | 7957 | 0.996 | 0.000 | 0.000 | 0.000 |
| Attentive, 5 | fully attentive/highly engaged | 0 | 5 | 6 | 2674 | 2685 | 0.996 | 0.000 | 0.000 | 0.000 |

Table S6. Overall, Category-specific and Codebook-specific Performance of MOSAIC

| **Category / Codebook** | **Accuracy, % (95% CI)** | **Precision, % (95% CI)** | **Recall, % (95% CI)** | **F1, % (95% CI)** |
|---|---|---|---|---|
| Overall | 95.3 (94.8–95.8) | 96.1 (95.5–96.6) | 95.3 (94.8–95.8) | 95.6 (95.1–96.1) |
| **By Codebook** | | | | |
| WISER | 94.1 (93.3–94.8) | 96.5 (95.8–97.2) | 94.1 (93.3–94.8) | 95.2 (94.6–95.9) |
| Patient Behaviors | 96.3 (95.6–97.0) | 95.9 (95.1–96.6) | 96.3 (95.6–97.0) | 96.1 (95.4–96.7) |
| Bias | 96.5 (95.6–97.3) | 96.5 (94.6–97.9) | 96.5 (95.6–97.3) | 96.4 (95.0–97.5) |
| Intervention (5As) | 95.8 (94.0–97.5) | 95.8 (93.6–97.7) | 95.8 (94.0–97.5) | 95.8 (94.2–97.4) |
| Global | 92.6 (90.6–94.7) | 94.5 (92.0–96.7) | 92.6 (90.6–94.7) | 93.5 (91.6–95.3) |
| **By Category** | | | | |
| Rheumatology | 95.6 (95.1–96.1) | 96.3 (95.7–96.9) | 95.6 (95.1–96.1) | 95.9 (95.4–96.4) |
| OB/GYN | 93.6 (91.9–95.4) | 94.7 (92.7–96.4) | 93.6 (91.9–95.3) | 94.1 (92.6–95.5) |

Table S7. Results of OLS Regression Results for F1 Score by Codebooks and Total Number of Sentences.

Formula: F1~Codebook + Sentences per transcript

| Term | Coefficient | Std. Error | 95% CI | t | p |
|---|---|---|---|---|---|
| Intercept | 0.9572 | 0.0071 | [0.9432, 0.9711] | 134.3704 | <0.001 |
| Codebook: Global (vs Bias) | -0.0309 | 0.0090 | [-0.0485, -0.0133] | -3.4355 | <0.001 |
| Codebook: Intervention (vs Bias) | -0.0038 | 0.0128 | [-0.0289, 0.0213] | -0.2946 | 0.768 |
| Codebook: Patient Behavior (vs Bias) | -0.0059 | 0.0066 | [-0.0189, 0.0071] | -0.8901 | 0.373 |
| Codebook: Wiser (vs Bias) | -0.0144 | 0.0067 | [-0.0275, -0.0012] | -2.1437 | 0.032 |
| Sentences per transcript | 2.28e-05 | 1.67e-05 | [-9.86e-06, 5.55e-05] | 1.3689 | 0.171 |

Model fit: N = 120; $R^2$ = 0.1139; Adjusted $R^2$ = 0.0750; AIC = -527.9360; BIC = -511.2110

Table S8. Results of OLS Regression Results for F1 Score by Codebook and Total Number of Words.

Formula: weighted F1 ~ Codebook + words per transcript

| Term | Coefficient | Std. Error | 95% CI | t | p |
|---|---|---|---|---|---|
| Intercept | 0.9597 | 0.0055 | [0.9489, 0.9705] | 174.8970 | <0.001 |
| Codebook: Global (vs Bias) | -0.0310 | 0.0089 | [-0.0485, -0.0135] | -3.4704 | <0.001 |
| Codebook: Intervention (vs Bias) | -0.0047 | 0.0126 | [-0.0294, 0.0201] | -0.3719 | 0.710 |
| Codebook: Patient Behavior (vs Bias) | -0.0059 | 0.0065 | [-0.0187, 0.0069] | -0.9025 | 0.367 |
| Codebook: Wiser (vs Bias) | -0.0143 | 0.0066 | [-0.0272, -0.0014] | -2.1741 | 0.030 |
| Words per transcript | 2.03e-06 | 1.18e-06 | [-2.79e-07, 4.34e-06] | 1.7228 | 0.085 |

Model fit: N = 120; $R^2$ = 0.1222; Adjusted $R^2$ = 0.0837; AIC = -529.0640; BIC = -512.3390

Table S9. Results of OLS Regression Results for F1 Score by Codebook and Total Number of Speaker Turns.

Formula: weighted F1 ~ Codebook + speaker turns per transcript

| Term | Coefficient | Std. Error | 95% CI | t | p |
|---|---|---|---|---|---|
| Intercept | 0.9539 | 0.0072 | [0.9399, 0.9679] | 133.4124 | <0.001 |
| Codebook: Global (vs Bias) | -0.0313 | 0.0089 | [-0.0488, -0.0138] | -3.5129 | <0.001 |
| Codebook: Intervention (vs Bias) | -0.0003 | 0.0129 | [-0.0256, 0.0251] | -0.0214 | 0.983 |
| Codebook: Patient Behavior (vs Bias) | -0.0070 | 0.0066 | [-0.0199, 0.0059] | -1.0608 | 0.289 |
| Codebook: Wiser (vs Bias) | -0.0156 | 0.0067 | [-0.0287, -0.0025] | -2.3364 | 0.019 |
| Speaker turns per transcript | 3.75e-05 | 1.91e-05 | [4.26e-09, 7.49e-05] | 1.9602 | 0.050 |

Model fit: N = 120; $R^2$ = 0.1287; Adjusted $R^2$ = 0.0905; AIC = -529.9575; BIC = -513.2326

Table S10. Summary Across Models

| Model | Length coefficient | Length p | $R^2$ | Adjusted $R^2$ | AIC | BIC |
|---|---|---|---|---|---|---|
| Sentences per transcript | 2.28e-05 | 0.171 | 0.1139 | 0.0750 | -527.9360 | -511.2110 |
| Words per transcript | 2.03e-06 | 0.085 | 0.1222 | 0.0837 | -529.0640 | -512.3390 |
| Speaker turns per transcript | 3.75e-05 | 0.050 | 0.1287 | 0.0905 | -529.9575 | -513.2326 |

Table S11. Subgroup/ Department Performance Differences

| **Department** | **Accuracy** | **Precision** | **Recall** | **F1 score** |
|---|---|---|---|---|
| Overall (n = 120) | 0.959 (0.939-0.974) | 0.968 (0.949-0.983) | 0.959 (0.939-0.974) | 0.962 (0.940-0.976) |
| Rheumatology (n = 103) | 0.960 (0.942-0.974) | 0.968 (0.951-0.983) | 0.960 (0.942-0.974) | 0.965 (0.944-0.976) |
| OB/GYN (n = 17) | 0.941 (0.897-0.965) | 0.947 (0.934-0.970) | 0.941 (0.897-0.965) | 0.943 (0.922-0.968) |
| P value (Rheumatology vs. OB/GYN) | 0.048 | 0.06 | 0.048 | 0.019 |

Notes: [1] Values are reported as median (interquartile range). [2] P values were calculated using two-sided nonparametric tests comparing rheumatology and OB/GYN transcripts. [3] OB/GYN encounters showed lower median performance across all metrics; differences reached statistical significance for accuracy, recall, and F1 score, but not for precision.

Table S12. MOSAIC Setup Time and Runtime Summary

Setup-time benchmarks for major MOSAIC initialization components and Update Agent codebook re-indexing, measured across five repeated runs. Times are reported in seconds.

a. Main Table

Timing results for retriever initialization, example-library embedding and index construction, startup warm-up, and codebook re-indexing via the Update Agent. For codebook re-indexing, the reported value for each run is the median across the six original codebooks.

| **Setup component** | **Run 1 (s)** | **Run 2 (s)** | **Run 3 (s)** | **Run 4 (s)** | **Run 5 (s)** | **Median (s)** | **Mean (s)** | **Range (s)** |
|---|---|---|---|---|---|---|---|---|
| Retriever initialization | 5.69 | 4.62 | 4.84 | 6.60 | 4.53 | 4.84 | 5.26 | 4.53-6.60 |
| Example-library embedding and index construction | 7.70 | 7.93 | 7.58 | 7.34 | 7.74 | 7.70 | 7.66 | 7.34-7.93 |
| Startup warm-up | 3.55 | 3.17 | 3.14 | 3.12 | 3.13 | 3.14 | 3.22 | 3.12-3.55 |
| Codebook re-indexing via Update Agent, median across original codebooks | 0.114 | 0.100 | 0.107 | 0.095 | 0.093 | 0.100 | 0.102 | 0.093-0.114 |

b. Codebook Re-indexing Details (Per-codebook times from the five repeated runs)

Per-codebook timing results for Update Agent re-indexing of the six original codebooks, with the median time reported for each codebook across the five repeated runs.

| **Codebook** | **Run 1 (s)** | **Run 2 (s)** | **Run 3 (s)** | **Run 4 (s)** | **Run 5 (s)** | **Median (s)** |
|---|---:|---:|---:|---:|---:|---:|
| Global | 0.198 | 0.155 | 0.161 | 0.170 | 0.138 | 0.161 |
| Wiser | 0.086 | 0.075 | 0.085 | 0.076 | 0.075 | 0.076 |
| Intervention | 0.169 | 0.149 | 0.128 | 0.114 | 0.111 | 0.128 |
| Bias | 0.142 | 0.125 | 0.136 | 0.118 | 0.116 | 0.125 |
| SDOH & weight | 0.037 | 0.033 | 0.040 | 0.031 | 0.029 | 0.033 |
| Patient Behavior | 0.038 | 0.059 | 0.038 | 0.032 | 0.029 | 0.038 |

c. Codebook Re-indexing Details (Per-run aggregate summaries for codebook re-indexing)

Per-run aggregate summaries of Update Agent codebook re-indexing time across the six original codebooks, reported as the median and mean across codebooks, together with overall median, mean, and range across the five repeated runs.

| **Aggregate** | **Run 1 (s)** | **Run 2 (s)** | **Run 3 (s)** | **Run 4 (s)** | **Run 5 (s)** | **Median (s)** | **Mean (s)** | **Range (s)** |
|---|---|---|---|---|---|---|---|---|
| Median across six original codebooks | 0.114 | 0.100 | 0.107 | 0.095 | 0.093 | 0.100 | 0.102 | 0.093-0.114 |
| Mean across six original codebooks | 0.112 | 0.100 | 0.098 | 0.090 | 0.083 | 0.098 | 0.097 | 0.083-0.112 |

Table S13. API Usage and Cost Statistics

a. Aggregate Azure OpenAI usage for the full MOSAIC annotation workflow.

This table summarizes model-level request volume and token consumption for the full MOSAIC workflow across 174 transcripts (successful run of development and evaluation sets), which used two GPT-4o coders and one GPT-5.2 adjudicator. Azure monitoring metrics were aggregated separately for the coder and adjudicator deployments and then combined to estimate total workflow-level usage.

| **Model deployment** | **Total requests** | **Input tokens** | **Output tokens** | **Total tokens** |
|---|---|---|---|---|
| GPT-4o coder deployment | 10.03K | 16.87M | 5.00M | 21.87M |
| GPT-5.2 judger deployment | 6.99K | 25.14M | 5.77M | 30.92M |
| Combined | 17.02K | 42.01M | 10.77M | 52.79M |

b.Per-transcript Azure OpenAI usage and estimated API cost for the full MOSAIC workflow.

This table reports average request volume, token use, and estimated API cost per transcript, calculated by dividing aggregate Azure monitoring metrics by 174 transcripts. Cost estimates were calculated using public per-token pricing assumptions for GPT-4o and GPT-5.2 and do not account for cached-token discounts, institutional pricing, regional pricing differences, or Azure billing adjustments. Token-based billing depends on the selected model's input and output rates.

| **Metric** | **Per transcript** |
|---|---|
| Requests | 97.8 requests/transcript |
| Input tokens | 241,437 input tokens/transcript |
| Output tokens | 61,897 output tokens/transcript |
| Total tokens | 303,391 tokens/transcript |
| Estimated GPT-4o coder cost | $0.53/transcript |
| Estimated GPT-5.2 adjudicator cost | $0.72/transcript |
| Estimated total API cost | $1.25/transcript |

Table S14. Explainability and Error Analysis

a. Final-Output Correct Output Counts Overall, by Dataset Split, and by Coding Framework.

Counts of correctly classified final-output labels produced by MOSAIC after adjudication, summarized overall and stratified by dataset split and coding framework. "Correct output count" denotes the number of final-output labels that matched the gold-standard annotation.

| Section | Category | Correct output count |
|---|---|---|
| **Overall** | All frameworks | 60538 |
| **By split** | Development | 17339 |
| | Evaluation | 43199 |
| **By codebook** | Intervention | 1078 |
| | SDOH | 713 |
| | Patient Behavior | 20376 |
| | Bias | 13039 |
| | Global | 5171 |
| | Wiser | 20161 |

b. Final-Output Error Distribution Overall, by Dataset Split, and by Coding Framework.

Distribution of final-output errors after adjudication, summarized overall and stratified by dataset split and coding framework. Errors were grouped into unsupported or hallucinated labels, missed labels, label confusion, and other errors; total errors represent the sum across these categories.

| Section | Category | Unsupported or hallucinated | Missed label | Label confusion | Other | Total errors |
|---|---|---|---|---|---|---|
| **Overall** | All final-output errors | 1708 | 1159 | 27 | 8 | 2902 |
| **By split** | Development | 455 | 391 | 7 | 4 | 857 |
| | Evaluation | 1253 | 768 | 20 | 4 | 2045 |
| **By codebook** | Bias | 265 | 161 | 2 | 0 | 428 |
| | Global | 238 | 123 | 3 | 8 | 372 |
| | Intervention | 25 | 21 | 1 | 0 | 47 |
| | Patient Behavior | 349 | 439 | 2 | 0 | 790 |
| | SDOH | 0 | 43 | 0 | 0 | 43 |
| | WISER | 831 | 372 | 19 | 0 | 1222 |

Table S15. Disagreement Between Dual Coders and Review-queue Counts Overall, by Dataset Split, and by Coding Framework.

Counts of disagreement types between the two independent coders before final adjudication, summarized overall and stratified by dataset split and coding framework. Disagreements were categorized as alignment or format error, tag-versus-none disagreement, and multi-tag difference. The review-queue columns report the subset of disagreements flagged for human review after adjudication.

| Section | Category | Alignment or format error, all disagreements | Tag vs none, all disagreements | Multi-tag difference, all disagreements | Alignment or format error, review queue | Tag vs none, review queue | Multi-tag difference, review queue |
|---|---|---|---|---|---|---|---|
| **Overall** | All transcripts | 1722 | 1509 | 641 | 1722 | 772 | 519 |
| **By split** | Development | 640 | 351 | 168 | 640 | 221 | 151 |
| | Evaluation | 1082 | 1158 | 473 | 1082 | 551 | 368 |
| **By codebook** | Bias | 192 | 211 | 12 | 192 | 31 | 2 |
| | Global | 438 | 232 | 102 | 438 | 17 | 67 |
| | Intervention | 181 | 27 | 6 | 181 | 4 | 3 |
| | Patient Behavior | 363 | 686 | 432 | 363 | 681 | 432 |
| | SDOH | 398 | 2 | 0 | 398 | 2 | 0 |
| | WISER | 150 | 351 | 89 | 150 | 37 | 15 |

Table S16. Confidence and Rationale-Derived Ambiguity Signals

Confidence and rationale-derived ambiguity signals from disagreement and review-queue outputs, including low-confidence rates and the frequency of explanations citing missing rule support, scope limitations, or text-limited interpersonal cues. Confidence reflects the adjudication layer's estimated certainty that the selected label is adequately supported by the retrieved coding rules and the available transcript evidence.

a. Confidence Profile by Disagreement Type Before and After Review-queue Filtering

| Disagreement type | Dataset | N | Mean confidence | Median confidence | Proportion < 0.7 |
|---|---|---|---|---|---|
| **Alignment or format error** | All disagreements | 1722 | 0.0000 | 0.00 | 1.0000 |
| | Review queue | 1722 | 0.0000 | 0.00 | 1.0000 |
| **Multi-tag difference** | All disagreements | 641 | 0.6491 | 0.65 | 0.6022 |
| | Review queue | 519 | 0.6021 | 0.62 | 0.7437 |
| **Tag vs none** | All disagreements | 1509 | 0.6916 | 0.78 | 0.4248 |
| | Review queue | 772 | 0.5383 | 0.55 | 0.8303 |

b.Rationale-derived ambiguity flags in the review queue

| Rationale flag | Count |
|---|---|
| No rule support | 949 |
| Text-only tone limit | 80 |
| Scope or holistic judgment | 48 |

Table S17. Representative Explainability and Failure-Mode Examples

| Example type | Codebook | Transcript snippet | Gold / coder context | MOSAIC output or adjudication | Confidence | Interpretation |
|---|---|---|---|---|---|---|
| **Explainability success** | Wiser | “Oh, nice. When was that?” | Coder 1: Wiser: OE; Coder 2: None | Adjudicated: Wiser: OE | 0.91 | The rationale linked the utterance to the retrieved open-ended-question rule, illustrating traceable rule-to-label alignment. |
| **Hallucinated label** | Global | “Doctor: yeah” | Gold: None | Predicted: Attentive: 4 | NA | A minimal acknowledgment was over-interpreted as evidence of sustained attentiveness. |
| **Missed label** | Global | “Doctor: Okay” | Gold: Attentive, 5 | Predicted: None | NA | A potentially meaningful local response was left unlabeled, consistent with omission errors in holistic Global coding. |
| **Label confusion** | Bias | “What are you going to do about that?” | Gold: S | Predicted: J | NA | The system identified a bias-related signal but selected the wrong label. |
| **Low-confidence triage** | Patient Behavior | “No they want it to be on 50.” | Coder 1: None; Coder 2: Patient Behavior: AR | Adjudicated: None | 0.20 | The rationale explicitly cited missing rule support, showing how the review queue surfaces weakly grounded cases for human review. |
| **Global text-only limitation** | Global | “Yeah, we're not like a road trip destination, you know?” | Coder 1: Warmth: 4; Coder 2: Global: Warmth: 4 | Adjudicated: None | 0.62 | The rationale noted that an isolated humorous aside was insufficient for an encounter-level warmth judgment, highlighting the limits of transcript-only input. |

## Supplementary Figures

Figure S1. Internal ADS Tool and Transcript Generation.

This diagram summarizes the preprocessing workflow used to transform clinical audio recordings into structured transcripts for downstream MOSAIC annotation. The original version of our internally developed ADS system has been described previously.[1] In this study, we used the system as the foundation and further refined its components to enhance robustness. The pipeline integrates voice activity detection (VAD), audio-to-text transcription, speaker diarization, and LLM-based structured note generation. During VAD, clinician–patient speech intervals were identified,[2] using a combination of energy-based detection,[3] Silero VAD,[4] and SpeechBrain,[5] with majority voting to extract accurate segments. Clinical audio was transcribed with OpenAI Whisper-turbo.[6] Speaker diarization relied on a voice-embedding and clustering approach via Pyannote,[7] with detected segments aligned to Whisper transcriptions at sub-second resolution. Speaker role assignment (clinician vs. patient) was further processed based on the reasoning ability of GPT-4o[8] with the self-consistency prompting,[9] where the model was independently prompted 10 times using representative utterances for each speaker. The final speaker roles would then be determined via majority voting across the 10 predictions. This process enabled the generation of structured transcripts that include timestamps (1 min per block), speaker roles, utterances, and silence notations. All processing of our internal system was conducted within either a secure institutional virtual machine environment or local environment to ensure clinical data security and patient privacy. Personally identifiable information was removed or masked, ensuring compliance with standard clinical data governance practices. After preprocessing, the structured transcripts are ready for the LangGraph-based[10] MOSAIC coding. This diagram was created using diagrams.net.

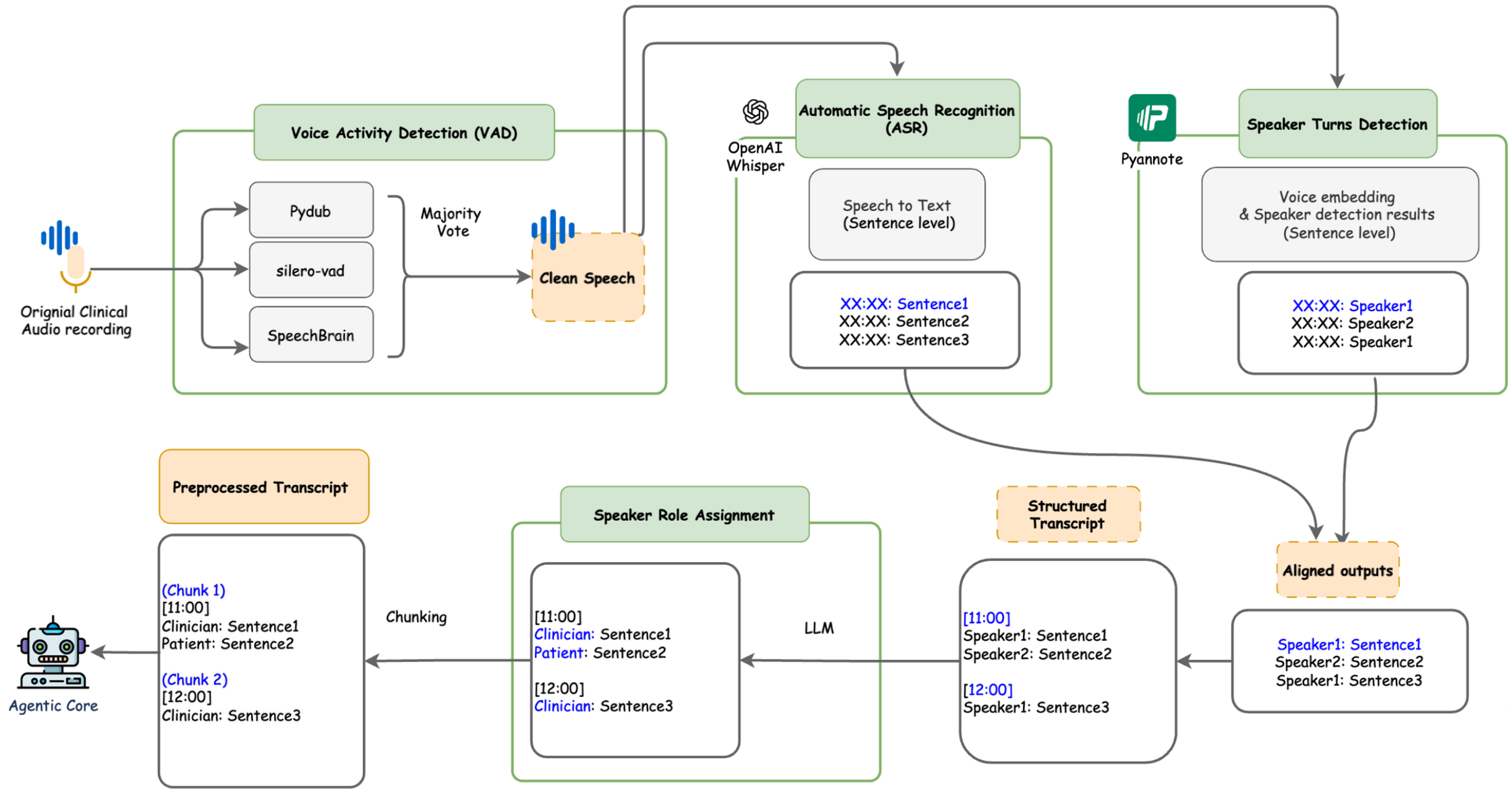

Figure S2. RAG Techniques in the Annotation Agent.

This diagram illustrates how MOSAIC retrieves clinician-authored coding rules during annotation. All text-based codebook rules and transcript segments are embedded using all-MiniLM-L6-v2.[11] Each transcript segment is used as a retrieval query against a FAISS-indexed codebook vector store.[12] The system retrieves candidate codebook rules, applies maximum marginal relevance to balance relevance and diversity, reranks transcript–rule pairs using a MedCPT cross-encoder,[13] deduplicates redundant rules, and injects the highest-ranked rules into the final annotation prompt. This retrieval and reranking process provides framework-specific rule context for the Annotation Agent and helps constrain outputs to valid codebook labels. This diagram was created using diagrams.net.

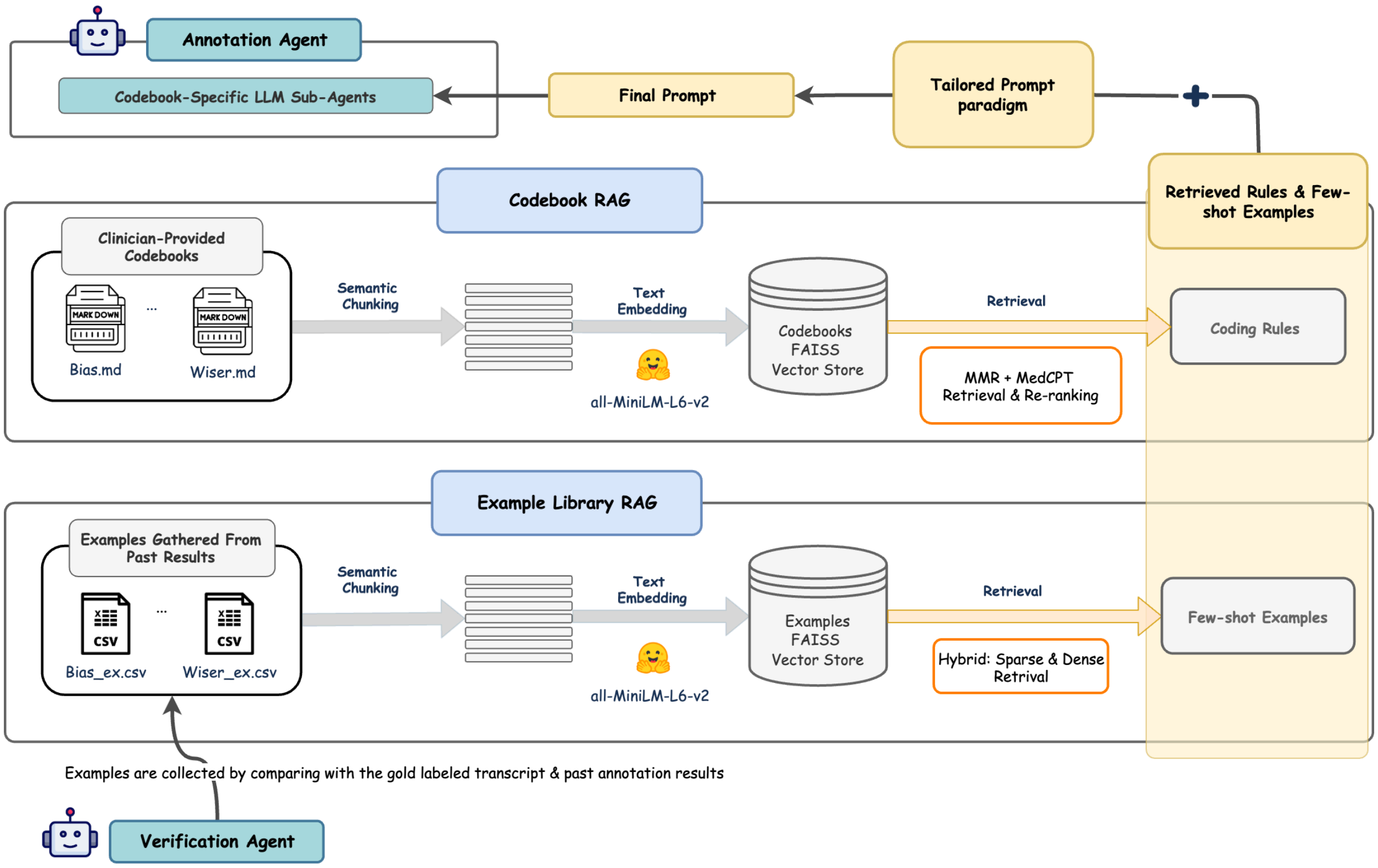

Figure S3. Example Library RAG Techniques in Dynamic Prompting Strategy.

This diagram summarizes how MOSAIC uses verification-derived examples to support dynamic in-context prompting. Sentence-level examples are paired with local context and labeled according to prior verification outcomes, including false negatives, false positives, true positives, and true negatives. During later annotation runs, examples are retrieved using a hybrid BM25 and FAISS strategy. False-negative examples are prioritized as Focus cases, indicating labels that should be assigned, whereas false-positive examples are prioritized as Avoid cases, indicating labels that should not be assigned. Retrieved examples are injected into the annotation prompt to guide subsequent coding behavior. This diagram was created using diagrams.net.

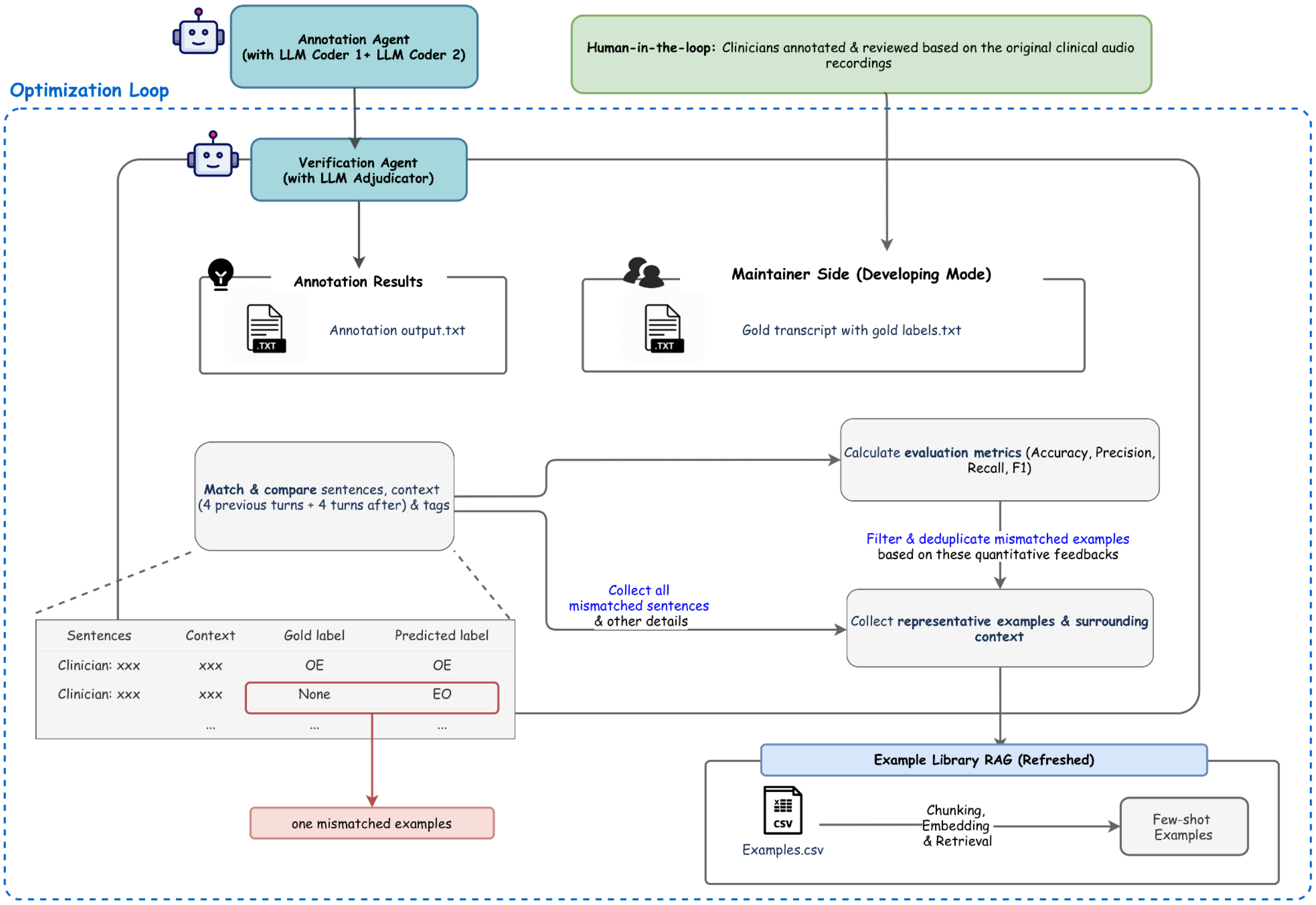

Figure S4. Input and Output Structure (Synthetic).

This synthetic example illustrates the expected transcript input and MOSAIC annotation output format. Input transcripts preserve timestamped utterances and speaker roles. MOSAIC outputs preserve the original transcript order and content, appending framework-specific labels only when the corresponding codebook rules are satisfied. This structure supports sentence- and segment-level annotation while minimizing unintended text modification, paraphrasing, or hallucinated labels. This diagram was created using diagrams.net.

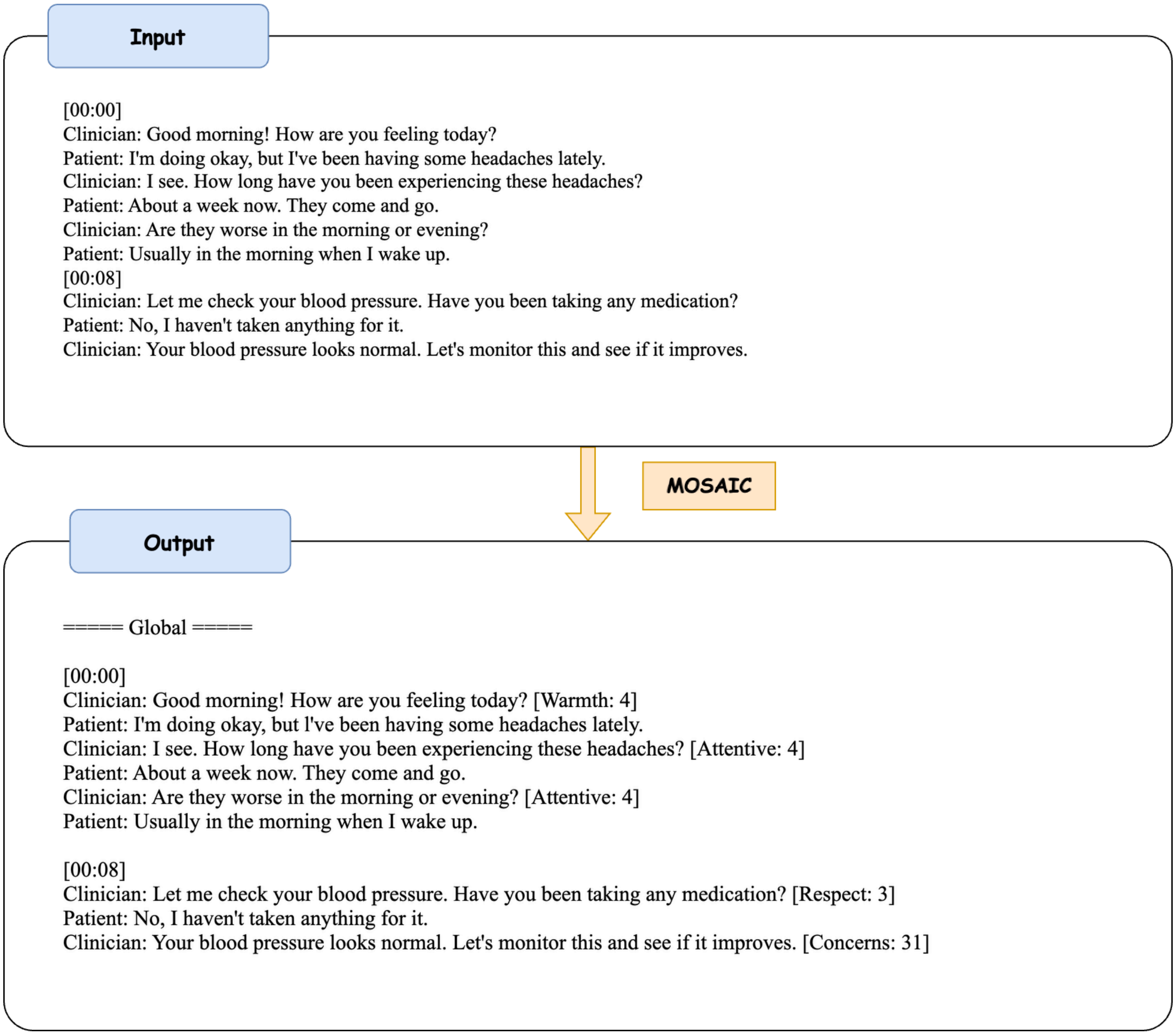

Figure S5. Markdown-Based Codebook Structure and Example Library Representation.

This diagram shows how clinician-authored codebooks and verified examples are represented for retrieval. Codebooks are reformatted using Markdown headers and segmented into rule-level units before embedding and indexing. Verified examples are stored with sentence-level text, local context, labels, prediction outcomes, and verification-derived metadata. This structure supports both codebook-rule retrieval and dynamic example retrieval during annotation. This diagram was created using diagrams.net.

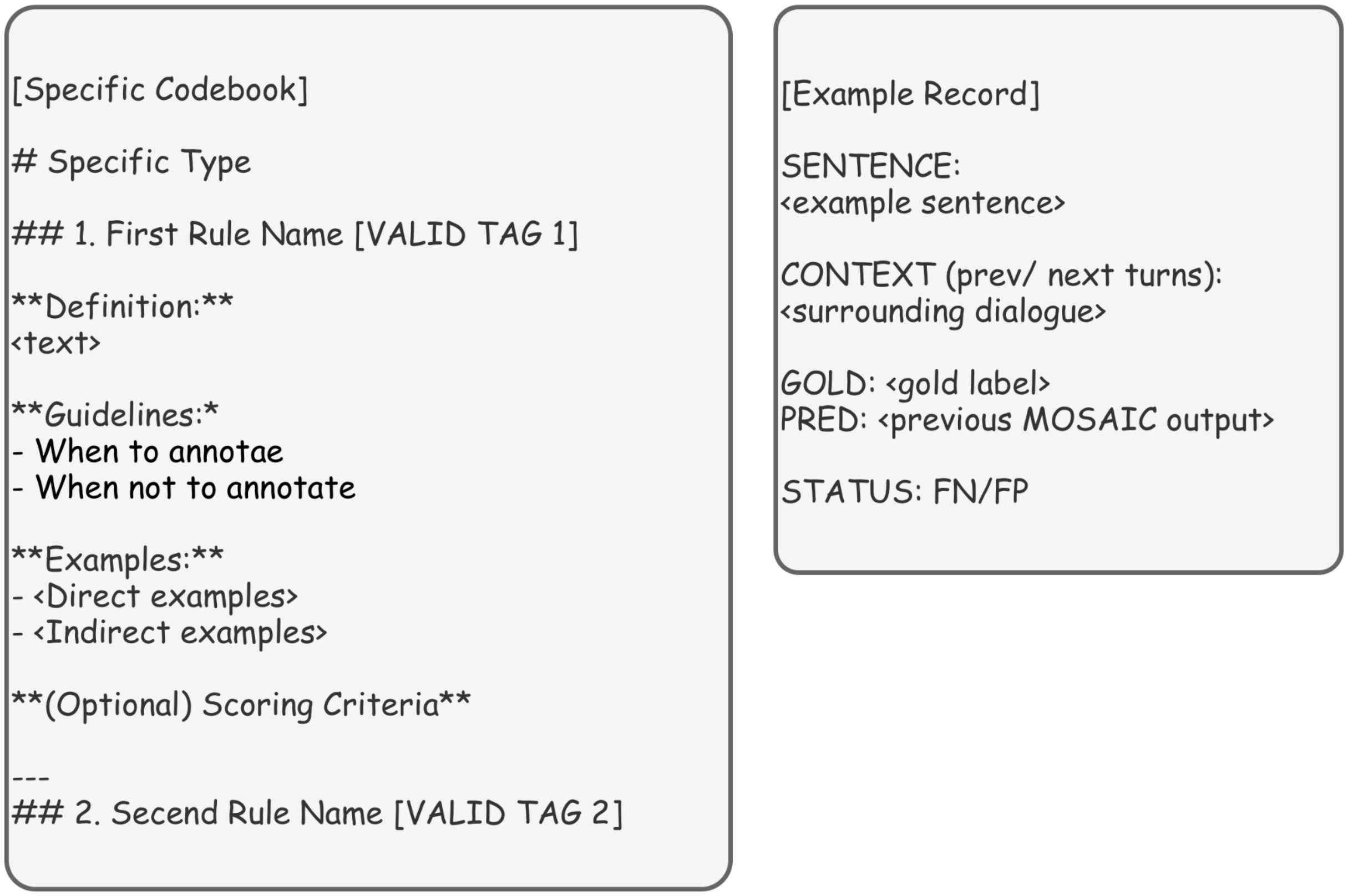

Figure S6. Modular Prompt Structure.

This diagram summarizes the structured prompt used by the Annotation Agent. For each transcript batch and requested framework, the final prompt consists of six components: agent role definition and global annotation instructions, framework-specific constraints, a whitelist of valid labels and formatting rules, retrieved and reranked codebook rules, dynamically selected Focus and Avoid few-shot examples, and the transcript batch to be annotated. This modular structure constrains outputs to clinician-defined labels, supports reproducible annotation behavior, and reduces unsupported labels, paraphrasing, and unintended text modification. This diagram was created using diagrams.net.

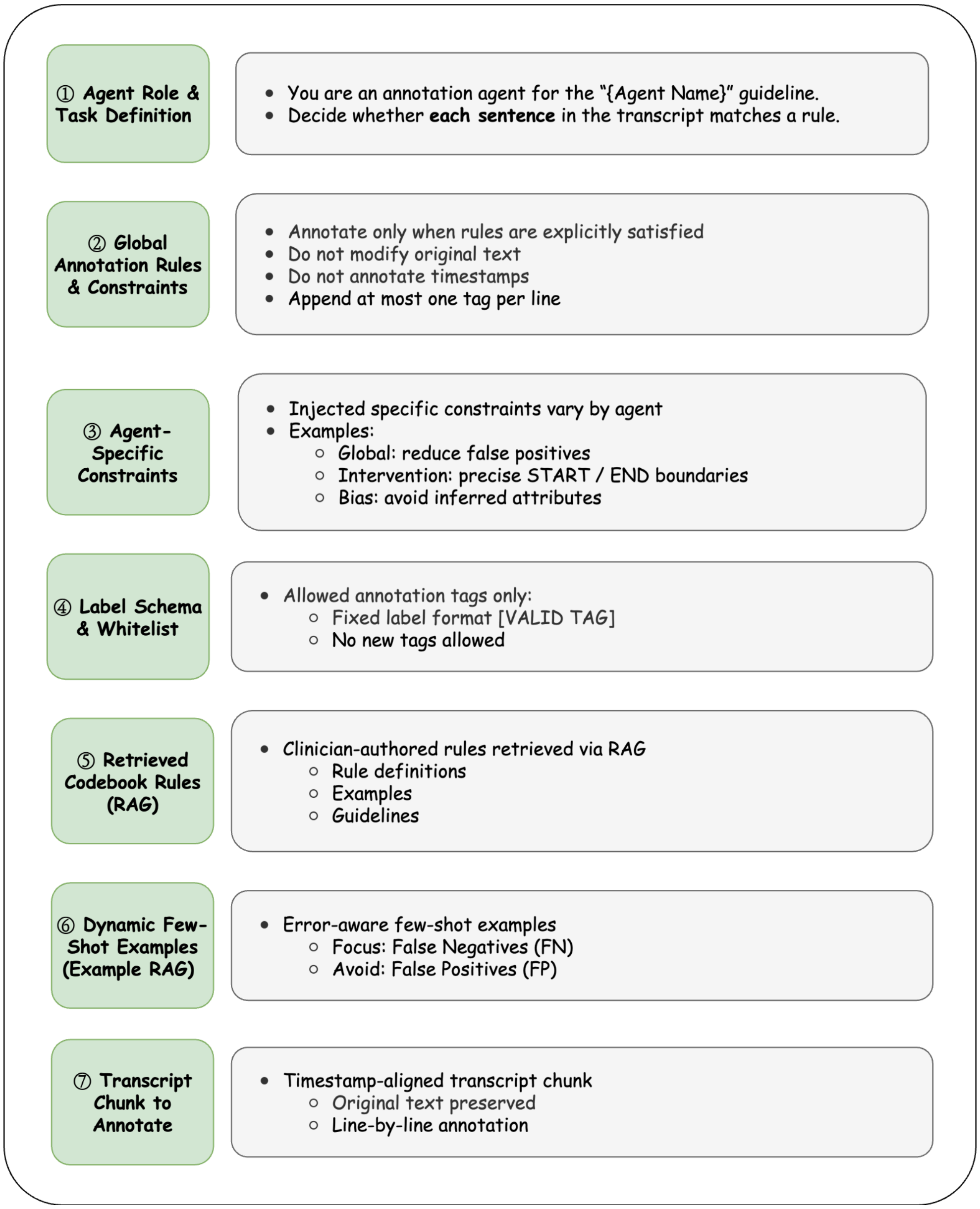

Figure S7. MOSAIC User Interface.

MOSAIC provides a web-based graphical interface designed to support transcript annotation, adjudication-aware verification, and report review for clinical communication analysis.

a. The annotation panel.

The Annotation Panel allows users to upload a transcript, enter an optional task prompt, select one or more framework-specific annotation agents (e.g., WISER, Bias, Intervention, Global, Patient Behavior, and SDOH & weight), optionally upload an updated codebook, and choose the adjudication model configuration. Uploaded transcripts are processed through the selected frameworks, and the interface returns the final adjudicated annotation output for preview, editing, and download in text or document format. This screenshot was captured from the MOSAIC web-based interface.

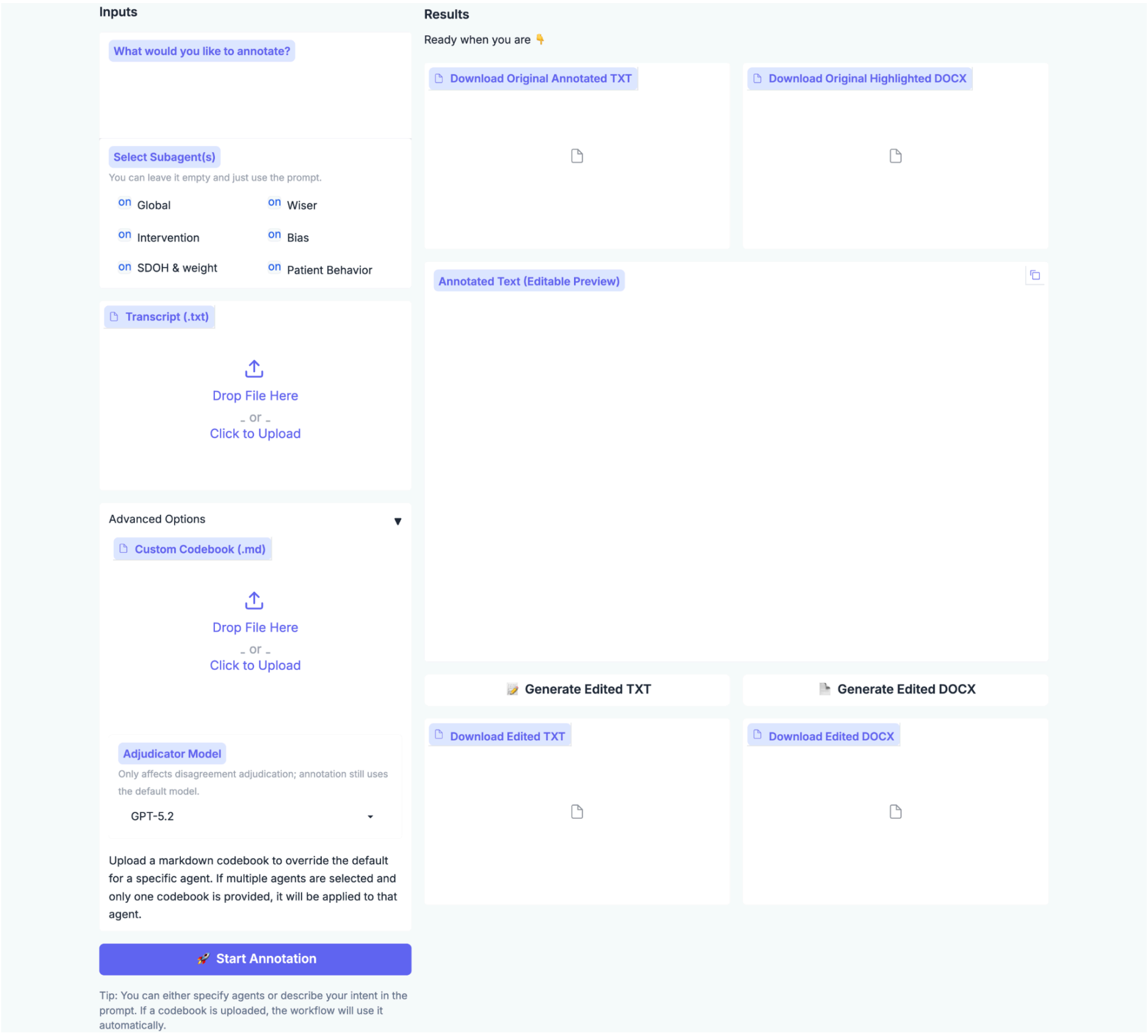

b. The verification panel.

The Verification Panel supports quality assessment when gold-standard reference transcripts are available. Users may upload a gold-labeled transcript to trigger gold-label verification, after which the interface displays computed evaluation metrics, along with matched outputs, mismatch reports, and per-label summaries. In addition to quantitative metrics, the interface presents retrieved rule snippets, LLM-generated summary and narrative explanations, and a structured human review queue for low-confidence or ambiguous adjudication cases. Users can preview report tables directly in the browser and download the full matched, mismatched, overall-metric, per-label, and review-queue files for further analysis. This screenshot was captured from the MOSAIC web-based interface.

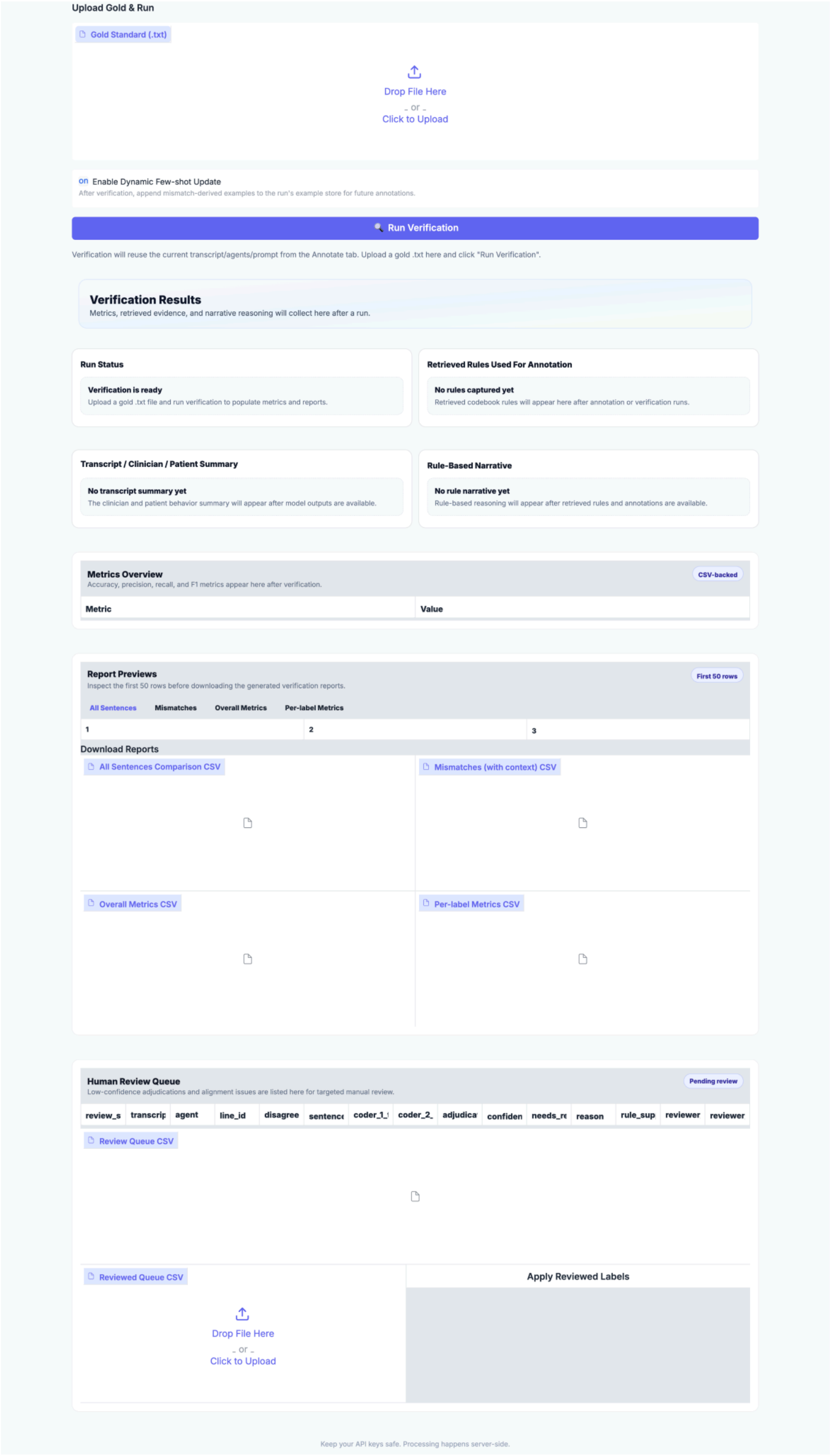

Figure S8. The Comparison of Different Regression Analysis on F1 vs. Text length.

Each panel shows transcript-level F1 scores plotted against a transcript-length measure: sentences per transcript, words per transcript, or speaker turns per transcript. Points represent transcript-level evaluation results, and fitted trend lines summarize the linear association between each length measure and F1. Pearson correlation, Spearman correlation, and sample size are shown within each panel. The weak correlations suggest that MOSAIC performance is not strongly associated with transcript length. This figure was created using Python.

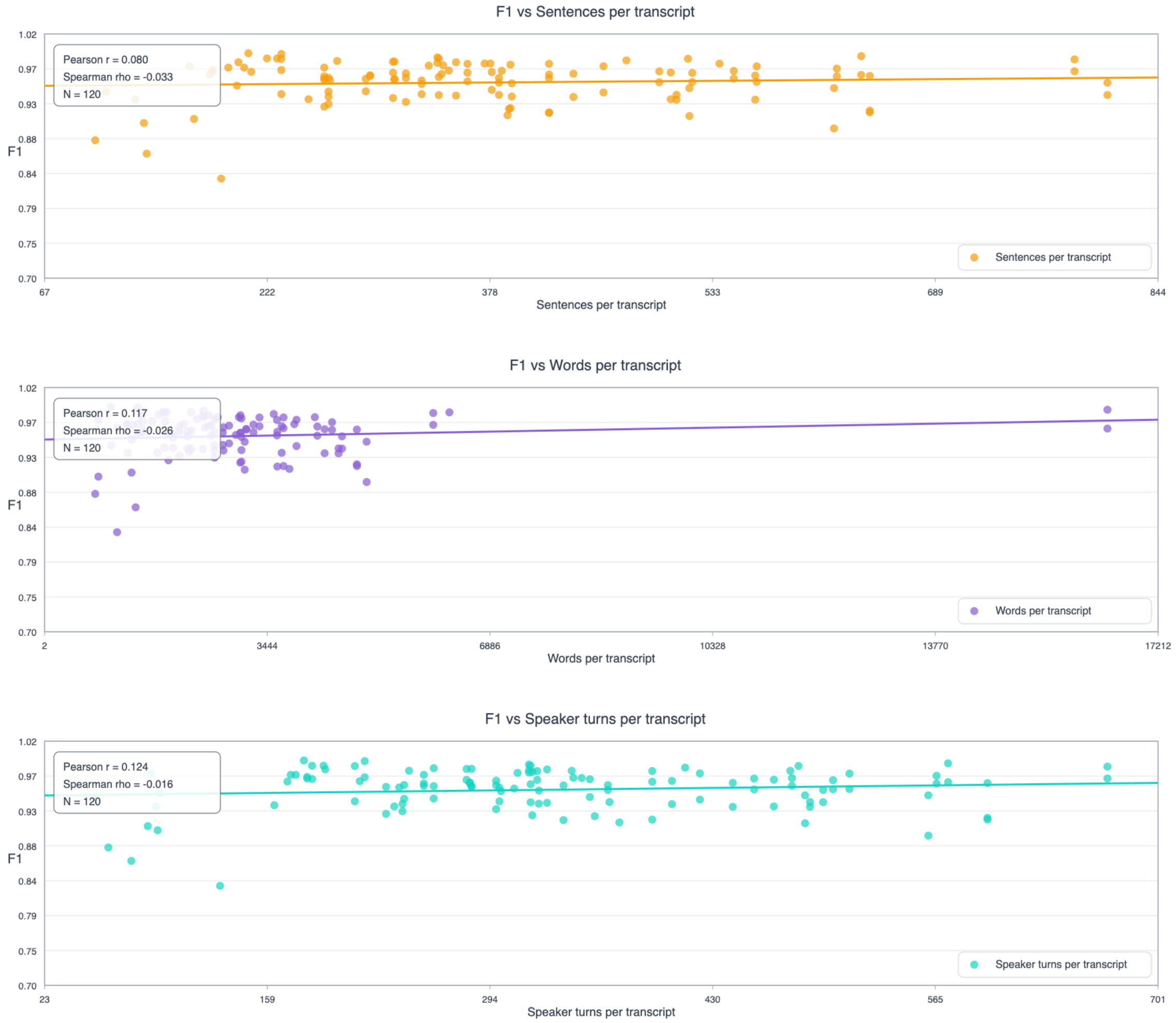